\documentclass{article}

\PassOptionsToPackage{numbers,sort&compress}{natbib}

\usepackage[preprint]{neurips_2026}

\usepackage{microtype}
\usepackage{graphicx}
\usepackage{subcaption}
\usepackage{booktabs} 
\usepackage{bbm}
\PassOptionsToPackage{hyphens}{url}
\usepackage{hyperref}
\usepackage[nobiblatex]{xurl}

\usepackage{amsmath}
\usepackage{amssymb}
\usepackage{mathtools}
\usepackage{amsthm}

\usepackage{enumitem}
\usepackage[capitalize,noabbrev]{cleveref}

\usepackage[most]{tcolorbox}
\usepackage{xcolor}
\usepackage{xspace}
\usepackage{wrapfig}

\newtcolorbox{myanswerbox}{
    colback=gray!15,       
    colframe=gray!15,      
    arc=5pt,               
    boxrule=0pt,           
    left=2pt,             
    right=2pt,            
    top=2pt,               
    bottom=2pt,            
    fontupper=\itshape,    
    halign=justify         
}

\theoremstyle{plain}

\theoremstyle{definition}

\theoremstyle{remark}

\usepackage[textsize=tiny]{todonotes}

\newcommand{\minihead}[1]{\vspace{0.1em}\noindent\textbf{#1}}

\newcommand{\eg}{e.g.}
\newcommand{\ie}{i.e.}

\newcommand{\releaseinfo}{We have submitted our curated datasets and code in the 
Supplementary Material.}

\title{Noisy Data is Destructive to Reinforcement Learning with Verifiable Rewards}

\author{%
  Yuxuan Zhu \\
  UIUC \\
  \texttt{yxx404@illinois.edu} \\
  \And
  Daniel Kang \\
  UIUC \\
  Bridgewater AIA Labs \\
  \texttt{ddkang@illinois.edu}
}

\begin{document}

\maketitle

\begin{abstract}
Reinforcement learning with verifiable rewards (RLVR) has driven recent 
capability advances of large language models across domains. Recent 
studies suggest that improved RLVR algorithms allow models to learn 
effectively from incorrect annotations, achieving performance comparable to 
learning from clean data. In this work, we show that these findings are invalid
because the claimed 100\% noisy training data is ``contaminated'' with clean 
data. After rectifying the dataset with a rigorous re-verification pipeline, we 
demonstrate that noisy data is destructive to RLVR. We show that existing RLVR 
algorithm improvements fail to mitigate the impact of noisy data, achieving 
similar performance to that of the basic GRPO. Furthermore, we find that the 
model trained on truly incorrect annotations performs 8--10\% worse than the 
model trained on clean data across mathematical reasoning benchmarks. Finally, 
we show that these findings hold for real-world noise in Text2SQL tasks, where 
training on real-world, human annotation errors cause 5--12\% lower accuracy 
than clean data. Our results show that current RLVR methods cannot yet 
compensate for poor data quality. High-quality data remains essential.

\end{abstract}

\section{Introduction}

Reinforcement learning with verifiable rewards (RLVR) is a widely used 
post-training paradigm \cite{guo2025deepseek,openaio3}, improving the
reasoning capabilities of large language models (LLMs) \cite{shao2024deepseekmath,
deepswe2025,deepcoder2025,deepscaler2025,wei2025swe,wen2025reinforcement,
ma2025general,su2025crossing}. However, obtaining verifiable annotations at 
scale for reward calculation is labor-intensive, and ensuring their quality is 
difficult \cite{wretblad2024understanding,chen2025minimax,team2025kimi}. Recent 
studies suggest that data quality is secondary, arguing that 
models can learn equally effectively from cheaper, noisy data, leveraging
algorithmic improvements such as reward shaping \cite{cai2025reinforcement}, 
objective correction \cite{mansouri2025noise}, and clipping mechanisms 
\cite{park2025clip}. This raises a fundamental question for the 
future of post-training: \textit{To what extent can RLVR, with 
algorithmic improvements, tolerate annotation noise?}

A growing body of recent literature suggests a promising, yet counter-intuitive 
answer: RLVR appears robust to data noise. Several studies 
claim that LLMs can learn effectively from low-quality data or even random noise, 
achieving performance comparable to models trained on clean data 
\cite{shao2025spurious,park2025clip,lv2025climb}. For instance, 
\citet{shao2025spurious} report that RLVR with 100\% incorrect data annotations 
achieves accuracy on MATH-500 within 5\% of the same model trained on clean 
data. Similarly, \citet{lv2025climb} suggest that clipped objectives enable 
learning from random rewards with only 3\% performance loss. These findings 
increasingly imply that data quality is secondary to algorithmic design.

In this work, we argue that these findings are invalid by showing that the noisy 
data used in prior work is not 100\% noisy. Through a rigorous empirical 
study, we demonstrate that high-quality data remains necessary and cannot be 
replaced by existing algorithmic improvements.

\minihead{Re-verifying data reveals severe impact of noise.}
We identify a critical contamination issue in the training datasets used in prior
work. Using a data re-verification pipeline leveraging GPT-5 Pro and 
manual verification (Section \ref{sec:data}), we find that at least 16\% of annotations  labeled as ``incorrect''
\begin{wrapfigure}{r}{0.45\textwidth}
    \centering
    \includegraphics[width=\linewidth]{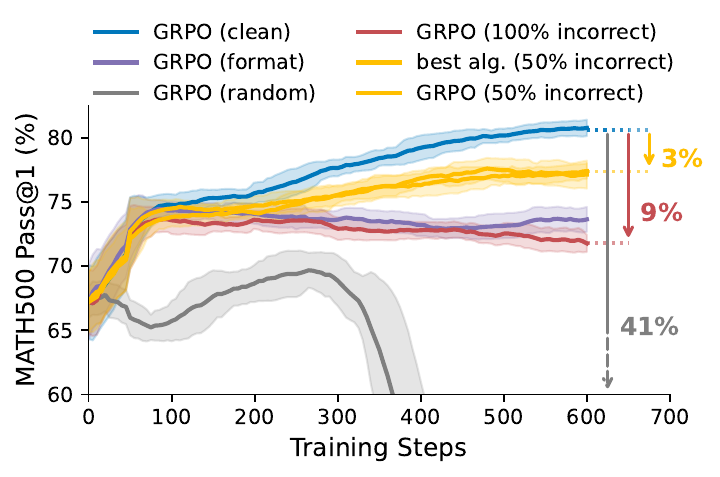}
    \caption{Noisy data significantly degrades RLVR performance, while existing 
    algorithms fail to mitigate it. Training Qwen2.5-Math-7B on 100\% incorrect 
    annotations performs similarly to format-only rewards and 9\% worse than 
    clean data. Even the best-performing RLVR algorithmic improvements 
    \cite{liu2025understanding,yao2025offpolicy,cai2025reinforcement,gao2025soft,
    yu2025dapo} trained on 50\% noise only match baseline GRPO 
    \cite{shao2024deepseekmath} and underperform the model trained on clean data.}
    \label{fig:summary}
\end{wrapfigure}
 in prior work are actually correct
(Figure \ref{fig:data-pipeline}). 
This contamination inflated the performance of models trained on noise. 
After removing the correct annotations from the prior dataset, we find 
that training on 100\% incorrect annotations degrades MATH-500 accuracy by 9\% 
(8-10\% on other benchmarks) compared to clean data (Figure \ref{fig:summary}). 
This performance is similar to training with format-only rewards (rewarding 
\texttt{$\backslash$}{boxed\{\}}), empirically showing that RLVR with pure noise 
fails to boost reasoning. Furthermore, we show that training on pure random 
annotations results in performance lower than the base model 
(Figure \ref{fig:summary}).

\minihead{Existing algorithms fail to compensate for noisy data.}
We evaluate state-of-the-art algorithmic improvements, including bias mitigation 
(Dr.\xspace GRPO \cite{liu2025understanding}, TIS \cite{yao2025offpolicy}, and 
PGFC \cite{cai2025reinforcement}), adaptive clipping (SAPO \cite{gao2025soft}), 
and dynamic sampling (DAPO \cite{yu2025dapo}), on noisy data. Contrary to prior 
work, we find that these carefully designed algorithms fail to compensate for 
true data noise. Under 50\% synthetic noise, these algorithms perform similarly 
to GRPO \cite{shao2024deepseekmath} and underperform GRPO with clean data by 
over 3\% on MATH-500 (Figure \ref{fig:summary}).


\minihead{Real-world annotation noise is destructive to RLVR.}
Moving beyond synthetic noise, we use Text2SQL (\ie, translating natural 
language questions to SQL queries) tasks as a case study to illustrate the 
impact of real-world annotation errors. Text2SQL datasets contain naturally 
occurring human annotation errors due to the ambiguity of natural language and the 
multidimensionality of data. We manually corrected a Text2SQL dataset, BIRD 
\cite{li2023can}, which is known to contain substantial annotation noise 
\cite{wretblad2024understanding,liu2025nl2sql,pourreza2023evaluating}. We find 
that training on the real-world, noisy dataset degrades performance by 5–12\% 
compared to training on our corrected dataset across five base models.

We summarize our contributions as follows:
\vspace{-0.3em}
\begin{enumerate}[leftmargin=*,itemsep=1pt]
    \item We construct a truly noisy dataset of mathematical reasoning for RL 
    training. This provides the community with a reliable dataset to rigorously 
    investigate the impact of data noise in RLVR.
    \item We show that existing algorithmic improvements for RLVR fail to 
    mitigate the impact of noise. With 50\% noise, the best-performing algorithm 
    achieves similar accuracy to that of GRPO and 3\% lower than clean data.
    \item We demonstrate that the impact of data noise is more severe than 
    previously reported. Training with 100\% noise leads to similar or lower 
    performance than training with format rewards and 8-10\% lower than 
    clean data.
    \item Using Text2SQL tasks, we show that the presence of real-world 
    noise is destructive to RLVR. Compared to a cleaned dataset, the real-world, 
    noisy dataset lowers performance by 6--12\%.
\end{enumerate}

\section{Background}

\minihead{Reinforcement learning with verifiable rewards.}
RLVR has emerged as a dominant paradigm for post-training LLMs, with recent 
large-scale successes achieving state-of-the-art results in various domains 
\cite{guo2025deepseek,openaio3}. Group relative policy optimization (GRPO) is 
the first algorithmic instantiation of RLVR \cite{shao2024deepseekmath}. In each 
iteration, for a question-answer pair ($q, a$), GRPO samples a group of outputs 
$\{\tau_1, \ldots, \tau_G\}$ from the model $\pi_{\theta_{old}}$ and estimates 
the advantage of output $\tau_i$ as follows:
\begin{equation}
    \hat A_{i} = \frac{R(\tau_i, a) - \mathrm{mean}\left(\left\{R(\tau_j, a)\right\}_{j=1}^G\right)}{\mathrm{std}\left(\left\{R(\tau_j, a)\right\}_{j=1}^G\right)} 
    \label{eq:advantage}
\end{equation}
Then, GRPO optimizes the model by maximizing the following clipped 
objective:
\begin{equation*}
\begin{split}
    \mathbb{E}\left[ \frac{1}{G} \sum_{i=1}^G \frac{1}{|\tau_i|} \sum_{t=1}^{|\tau_i|} \left( \min \left( r_{i,t} \hat{A}_{i}, \text{clip}(r_{i,t}, 1\pm\epsilon) \hat{A}_{i} \right) \right) \right]
\end{split}
\end{equation*}

where $r_{i,t} = \pi_\theta(\tau_{i,t}|x, \tau_{i,<t}) / \pi_{\theta_{old}}(\tau_{i,t}|x, \tau_{i,<t})$.

As shown, advantage estimates $\hat A_{i}$ are the key optimization 
signals in RLVR, which relies on the quality of annotations $a$ (\eg, 
gold answers \cite{shao2024deepseekmath} and unit tests \cite{deepswe2025}). 
Since prior work assumes the availability of gold annotations 
\cite{shao2024deepseekmath,deepscaler2025,deepcoder2025,deepswe2025}, the impact 
of noise in $a$ remains underexplored.

\minihead{The noise robustness hypothesis.}
Recent literature argues that RLVR is robust to reward noise. 
\citet{shao2025spurious} observe that models trained with completely 
incorrect annotations ($a$) still improve on reasoning benchmarks significantly,
up to 24\% on MATH-500. They hypothesize that such reasoning improvements shown 
on benchmarks are due to the reasoning patterns learned in pre-training. 
Similarly, \citet{lv2025climb} suggest that RLVR improves LLM despite up to 40\% 
randomly flipped reward signals, achieving a substantial 67\% accuracy 
improvement on MATH-500. These findings increasingly imply the conclusion that 
RLVR has strong robustness to noise. Our findings challenge the validity of this 
robustness hypothesis. We demonstrate that the findings of 
\citet{shao2025spurious} are primarily because their noisy dataset is 
contaminated by correct annotations (Section \ref{sec:data}). Furthermore, we 
argue that \citet{lv2025climb} rely on a stochastic noise model that misaligns 
with the practical RLVR that typically uses deterministic, rule-based verifiers.

\minihead{Mitigation strategies for noisy rewards.}
To mitigate the biased optimization under noisy data, recent works have proposed 
various corrections to debias the reward signal. Under the assumption of 
independently and identically distributed (i.i.d.) reward noise, 
\citet{cai2025reinforcement} introduced Policy Gradient Backward Correction 
(PGBC) and Forward Correction (PGFC) to correct reward values given the noise 
rates. Similarly, \citet{mansouri2025noise} derived a corrected objective to 
debias the policy gradient. However, these methods rely on two strong 
assumptions: (1) the noise is i.i.d., and (2) noise rates can be accurately 
estimated. In practice, incorrect annotations are question-dependent and 
difficult to predict. We empirically show that these algorithmic mitigations 
fail to provide significant improvements under a question-dependent noise 
model (Section \ref{sec:algs}).

In addition, experimental evidence from \citet{shao2025spurious} and 
\citet{park2025clip} shows that the clipping mechanism can mitigate biased 
gradients caused by noisy data. In our study, with a truly noisy dataset, 
we show that even the advanced training stabilization using adaptive clipping 
\cite{gao2025soft} and dynamic sampling \cite{yu2025dapo} fail to outperform
the baseline GRPO (Section \ref{sec:algs}).

\section{Empirical Study Design}

In this section, we introduce the design of our empirical study, starting with 
three primary research questions that we aim to answer (\S \ref{sec:rqs}). 
Then, we discuss the types of noise we study (\S \ref{sec:noise-models}). 
Next, we present our method of training data curation for the empirical study 
(\S \ref{sec:data}). Finally, we introduce our implementation details of 
experiments (\S \ref{sec:impl}).

\subsection{Research Questions}
\label{sec:rqs}

To rigorously evaluate the limits of noise tolerance in RLVR, we must decouple 
data quality from confounding factors such as base models and algorithm variants. 
Therefore, we design a comprehensive study spanning \textit{six} 
base models, \textit{six} benchmarks, and \textit{five} algorithms besides GRPO. 
We begin by revisiting prior work's claims with a synthetic noisy dataset, 
before extending our analysis to real-world noise. Our study centers on 
three research questions:
\begin{enumerate}[label=RQ{\arabic*}.,leftmargin=2.7em,itemsep=1pt]
    \item Does the hypothesis that ``RLVR is robust to noise'' hold when 
    training data is truly noisy, without being contaminated 
    by correct annotations (Section \ref{sec:synthetic})?
    \item Can state-of-the-art algorithmic improvements (\eg, clipping, 
    bias mitigation) effectively recover the performance lost to severe data 
    noise (Section \ref{sec:algs})?
    \item Do findings from synthetic noise transfer to real-world data? 
    Specifically, how do natural human errors, such as those seen in
    Text2SQL datasets \cite{wretblad2024understanding}, impact RLVR performance
    compared with a clean dataset (Section \ref{sec:real-world})?
\end{enumerate}

\begin{figure*}
    \centering
    \includegraphics[width=\linewidth]{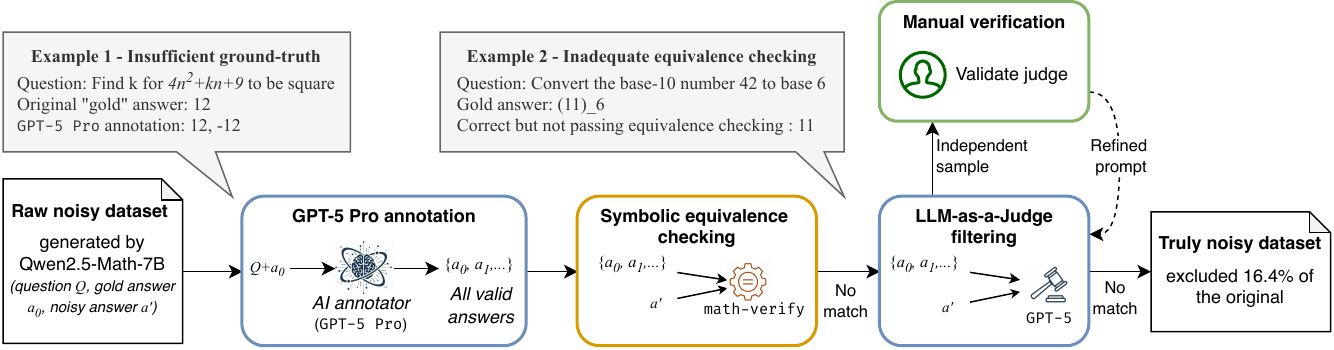}
    \caption{Data re-verification pipeline for synthesizing truly noisy dataset. 
    As shown in the two examples, we identify the issues of insufficient 
    ground-truth annotations and inadequate equivalence checking, and address 
    them via a rigorous pipeline combining LLMs and manual verification.}
    \label{fig:data-pipeline}
\end{figure*}

\subsection{Noise Models}
\label{sec:noise-models}

According to the RLVR reward formulation (Equation \eqref{eq:advantage}), 
an accurate reward signal relies on two components: a correct ground-truth 
annotation $a$ and a reliable equivalence verifier $R(a, \cdot)$. Consequently, 
noise in RLVR arises from two failures: (1) annotation errors ($a$ is incorrect), 
or (2) verifier errors (the function $R(\cdot, \cdot)$ is flawed). Based on this
categorization, we study four noise models as covered by prior work 
\cite{shao2025spurious,cai2025reinforcement,lv2025climb,mansouri2025noise}.

\minihead{Model-generated annotations.}
To simulate scalable, low-quality data, prior work generates ``noisy'' 
annotations using base models \cite{shao2025spurious}. We adopt this as our 
primary synthetic noise source. In addition, we apply a re-verification pipeline 
to ensure these annotations are genuinely incorrect, preventing the accidental 
inclusion of correct labels that invalidate the assumption.

\minihead{Random annotations.}
In addition to weak annotations, we study random annotations, a noise model 
widely used in prior work \cite{natarajan2013learning,zhang2018generalized}. For 
example, we generate a random integer 
$n \in [1, 1000]$ as the answer to a mathematical problem similar
to AIME problems \cite{aime}. This simulates a scenario where the data pipeline 
contains completely mismatched labels, serving as a lower-bound baseline for noise.

\minihead{Format reward.}
We use the format reward setting in prior work to simulate a collapsed verifier 
that checks only for the format of the LLM's generation \cite{shao2025spurious}. 
In this setting, a model receives a positive reward only for satisfying 
structural format requirements (\eg, enclosing an answer in 
\texttt{$\backslash$boxed\{\}}). We use this noise model to benchmark LLMs that 
learn format adherence without reasoning.

\minihead{Real-world noise.}
In practice, noise is often not purely random. It is systematic 
and ambiguous. Real-world datasets suffer from incorrect annotations, such as 
incorrect labels in ImageNet \cite{beyer2020we}, incorrect ``gold'' queries in 
BIRD \cite{wretblad2024understanding}, or insufficient test coverage in 
SWE-Bench \cite{yu2025utboost}. While verifier errors exist \cite{mathVerify}, 
they are often fixable engineering issues. In contrast, massive-scale annotation 
errors are a persistent bottleneck. Thus, the final part our study (\S 
\ref{sec:real-world}) investigates real-world annotation errors.


\subsection{Data Curation}
\label{sec:data}
Following the noise models, we curate datasets for both controlled, synthetic 
settings and real-world scenarios. For synthetic noisy data, we follow prior 
work \cite{shao2025spurious,lv2025climb,cai2025reinforcement,mansouri2025noise}
to use a mathematical reasoning dataset. For real-world noisy data, we use BIRD,
a Text2SQL dataset that is known to contain over 50\% annotation errors 
\cite{wretblad2024understanding,pourreza2023evaluating,jin2026pervasive,arcwise}.

\minihead{Synthetic noisy data.}
We used the same noisy dataset constructed by \citet{shao2025spurious}, 
which comprises incorrect generations of Qwen2.5-Math-7B on the DeepScaleR 
dataset \cite{deepscaler2025}. To inspect whether this noisy dataset has truly 
incorrect annotations, we investigated 20 randomly sampled examples and 
identified a high rate of invalid data. We found that 8/20 (40\%) of the 
examples labeled as ``incorrect'' were actually correct. There are two primary
causes for this data error:
\vspace{-0.5em} 
\begin{enumerate}[leftmargin=*,itemsep=1pt]
    \item \textit{Insufficient ground-truth annotation}: A math problem can have multiple 
    valid answers, while the ground-truth annotation only captures one of them 
    (Example 1 of Figure \ref{fig:data-pipeline}). Consequently, the
    dataset marks a correct annotation that differs from the ground-truth as 
    incorrect.
    \item \textit{Inadequate equivalence checking}: Weak symbolic equivalence verifiers 
    may fail to detect semantic equivalence across differing formats (Example 2 
    of Figure \ref{fig:data-pipeline}).
\end{enumerate}
To rigorously sanitize the noisy dataset, we propose the following multi-stage
data re-verification pipeline, targeting at reducing the number of correct 
annotations.
\vspace{-0.6em}
\begin{enumerate}[leftmargin=*,itemsep=1pt]
    \item \textit{GPT-5 Pro Annotation}: To address insufficient ground-truth 
    annotation, we used GPT-5 Pro to find all ground truth for each problem,
    given the problem text and the original ground-truth annotation.
    \item \textit{Symbolic Equivalence Checking}: We used \texttt{math-verify} 
    \cite{mathVerify} to check the target ``incorrect'' annotation against the 
    set of correct annotations.
    \item \textit{LLM-as-a-Judge Filtering}: To mitigate the issue of inadequate
    equivalence checking due to different formats, we applied GPT-5 judge to 
    evaluate the ``incorrect'' annotations that passed the symbolic check. We 
    prompted GPT-5 to determine if each ``incorrect'' annotation is equivalent 
    to any answer in the set of correct annotations. We attached the prompt in 
    Appendix \ref{sec:app-prompt}.
    \item \textit{Manual Investigation}: To validate the reliability of the LLM
    judge, we use an iterative refinement process. We first manually inspected 
    100 random samples. When we identified any LLM judge errors, we incorporated 
    the error patterns into the prompt and re-run the judging process with a new 
    independent sample. In our study, we conducted two iterations of refinement. 
\end{enumerate}

We justify the validity of using LLMs in the pipeline. First, 
while GPT-5 Pro may produce incorrect annotations, our objective is to remove 
correct annotations from the noisy dataset. Therefore, dropping a data point 
because it matches an incorrect annotation only reduces dataset size 
without contaminating the noise. Second, our manual investigation statistically
bounds the error rate of LLM judge. In the final sample of 100 
examples, we observed a 0\% error rate. Modeling this as a binomial 
proportion, we establish that the true error is upper-bounded at 
3\% with 95\% confidence \cite{newcombe1998two}.

In this pipeline, GPT-5 Pro identified 9.0\% of initial noisy dataset with 
insufficient ground-truth annotations. Together with inadequate equivalence 
checking, we excluded 16.4\% of the original noisy data, resulting in a set of 
12,769 mathematical problems with incorrect annotations. In addition to this 
noisy dataset, we construct a clean dataset and a random-annotation dataset.

\minihead{Real-world noisy data.}
We select Text2SQL tasks, translating natural language questions to SQL queries, 
as our target domain for curating real-world noisy dataset, since Text2SQL 
datasets inherently contain substantial annotation errors due to the semantic 
ambiguity of natural language and the data-dependent nature of writing SQL 
queries. Prior studies on Text2SQL datasets have shown that the widely used BIRD 
dataset \cite{li2023can} contain over 50\% incorrect annotations 
\cite{wretblad2024understanding,jin2026pervasive}. However, no existing work has 
curated a clean Text2SQL dataset to serve as a reliable reference for the impact 
of noise.

To quantify the impact of real-world noisy dataset on RLVR, we curated a 
clean dataset based on a random sample of 600 noisy Text2SQL instances from 
the BIRD Train set. For each Text2SQL instance, one of our authors proposed
correction while a second independently verified the correction. We repeated this
correction-verification pipeline until the verification passes. We resolved 
disagreement between correction and verification by involving another author.
This process identified and corrected 372 (62\%) noisy instances. We encountered
and resolved two correction-verification conflicts. In the rest of this paper, 
we denote the original 600 sampled BIRD Train instances as BIRD-600-Original and 
the corrected version as BIRD-600-Corrected.

\subsection{Implementation}
\label{sec:impl}
For relatively small-scale models (e.g., Qwen2.5-Math-7B), we used the SkyRL 
\cite{griggs2025skrylv01}. We conducted these experiments with four A100s, a 
64-Core CPU, and 512 GB RAM. For large-scale models (\eg, DeepSeek-V3.1), we 
used the Tinker, leveraging LoRA to achieve high efficiency 
\cite{tinker,schulman2025lora}. \releaseinfo

\section{Noisy Data Significantly Degrades RLVR Performance}
\label{sec:synthetic}

In this section, we investigate the impact of data quality on RLVR based on our
synthetic noisy dataset. We first describe the experimental settings and then 
present our findings.

\begin{figure*}[t!]
    \begin{subfigure}{0.33\textwidth}
        \includegraphics[width=\linewidth]{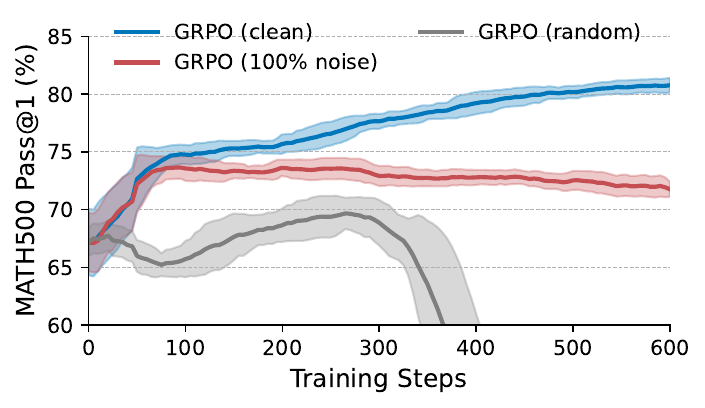}
    \end{subfigure}\hspace*{\fill}
    \begin{subfigure}{0.33\textwidth}
        \includegraphics[width=\linewidth]{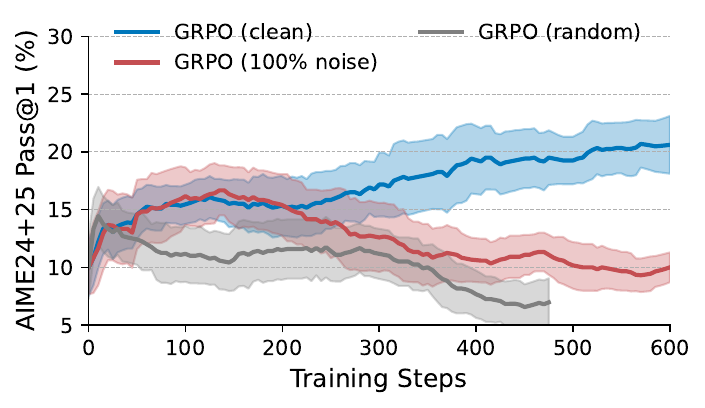}
    \end{subfigure}\hspace*{\fill}
    \begin{subfigure}{0.33\textwidth}
        \includegraphics[width=\linewidth]{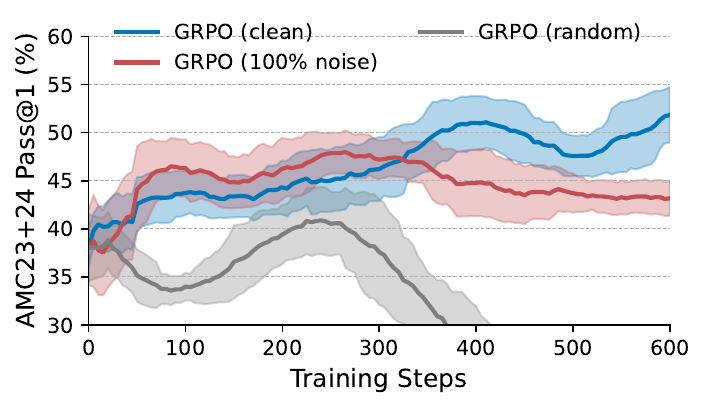}
    \end{subfigure}

    \begin{subfigure}{0.33\textwidth}
        \includegraphics[width=\linewidth]{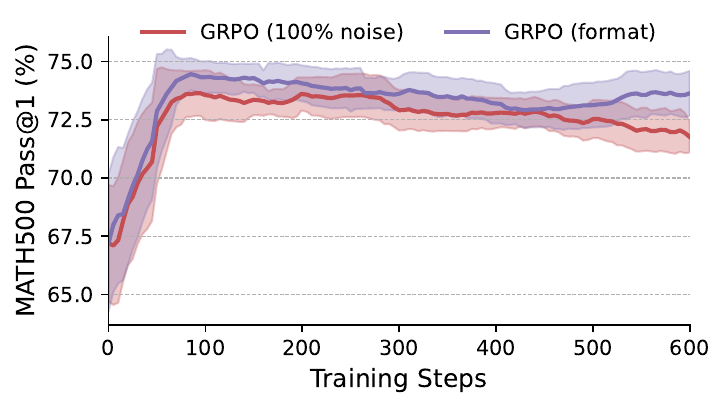}
    \end{subfigure}\hspace*{\fill}
    \begin{subfigure}{0.33\textwidth}
        \includegraphics[width=\linewidth]{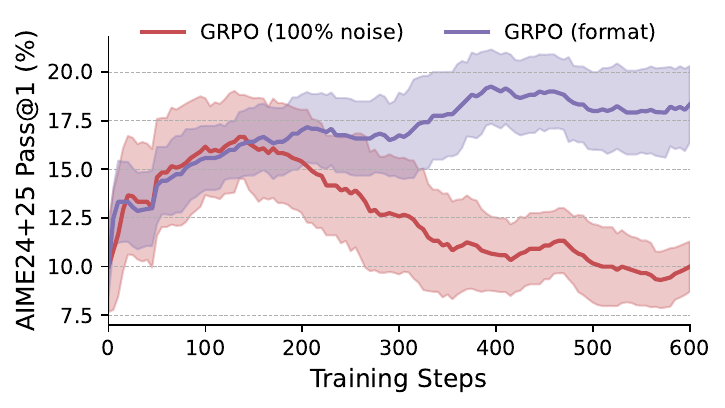}
    \end{subfigure}\hspace*{\fill}
    \begin{subfigure}{0.33\textwidth}
        \includegraphics[width=\linewidth]{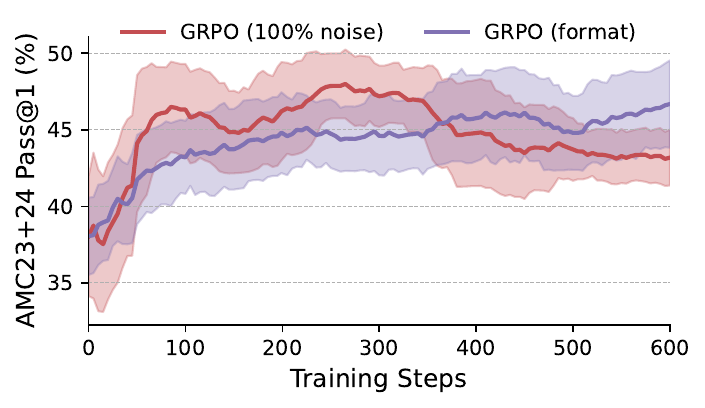}
    \end{subfigure}

        
    \caption{Both incorrect and random annotations significantly 
    decrease the performance of RLVR: RLVR with incorrect 
    annotations achieves comparable or lower accuracy (by 1.2--5.6\%) than RLVR with format rewards, and lower accuracy (by 8.5--10.0\%) than RLVR with clean data; RLVR with random annotations achieves significantly lower accuracy 
    (by 6.7--27.8\%) than the base model.} 
    \label{fig:synthetic}
\end{figure*}

\subsection{Experimental Settings}

\minihead{Training.}
We used Qwen2.5-Math-7B as our base model and trained it using the GRPO 
\cite{shao2024deepseekmath}. For reward computation, we used \texttt{math-verify} 
and assigned reward=1 if the generated answer is verified to match the 
solution and reward=0 otherwise. In the setting of format reward, we assign 
reward=1 if the generated sequence contains \texttt{$\backslash$boxed\{\}} and 
reward=0 otherwise. We set the batch size to 64, the group size to 16,
and the learning rate to $5\times 10^{-6}$, while keeping other settings 
(including the number of epochs) consistent with the configuration in 
\citet{shao2025spurious}. For the clean, noisy, and format-reward settings, we 
trained models for three epochs, as we observed no further improvements beyond 
this point (Appendix \ref{sec:app-ca}). For the random annotation setting, we 
used early stopping at two epochs as our pilot runs indicated significant 
performance collapse at the end of the second epoch.

\minihead{Evaluation.}
We performed all evaluations using greedy decoding (temperature=0). We include 
five benchmarks for evaluation: MATH-500 \cite{lightman2023let}, AIME 2024 and 
2025 \cite{aime}, AMC 2023 and 2024 \cite{amc}. Among them, the AIME 2025 and 
AMC 2024 datasets were released after the last update date of the model weights 
of our base model, ensuring a contamination-free evaluation environment. 

\subsection{Findings}

\begin{myanswerbox}
    \textbf{\upshape Answer to RQ1}. Noisy data impacts learning significantly, 
    resulting in 8-10\% lower accuracy than training on clean data and failing 
    to outperform training with pure format rewards.
\end{myanswerbox}

\minihead{Noisy data causes significant performance degradation.} We quantify 
the performance gap between training on clean data and training on noisy data. 
As shown in Figure \ref{fig:synthetic}, training on 100\% noise 
yields a substantially lower performance compared to the clean baseline, degrading accuracy by 9.0\% on MATH-500, 10.0\% on AIME, and 8.5\% on 
AMC. We also find that training on 100\% noise fails to outperform the 
format-reward baseline, suggesting that noisy data does not 
improve the capability of the base model. 
Furthermore, training on random annotations leads to catastrophic 
collapse, causing performance to drop significantly below that of the base model.

We attribute the accuracy degradation of the model trained on noisy data 
to a confirmation bias loop during optimization. As detailed in the training 
dynamics analysis in Appendix \ref{fig:app-dynamics}, we find that the model 
easily exploits its pre-existing flawed priors, achieving a 149\% higher average 
reward on noisy data while exhibiting a 9\% drop in policy entropy, rather than 
exploring correct reasoning paths.

\begin{figure*}[t!]
    \centering
    \begin{minipage}{0.32\linewidth}
        \centering
        \includegraphics[width=\linewidth]{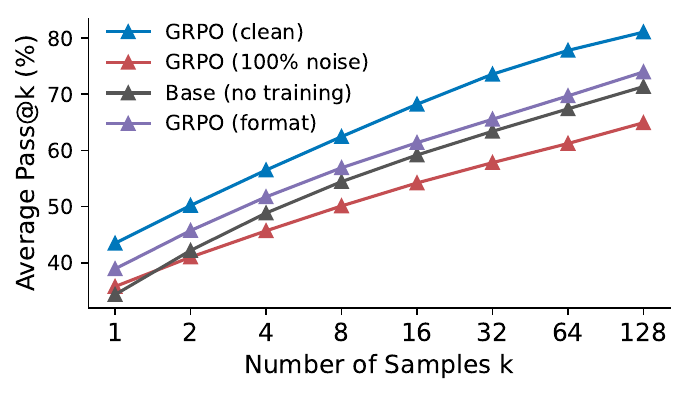}
        \caption{Noise leads to lower pass@$k$ ($k>1$) than base model, showing that noise fail to improve reasoning boundary.}
        \label{fig:pass_k_noise}
    \end{minipage}
    \hfill
    \begin{minipage}{0.32\linewidth}
        \centering
        \includegraphics[width=\linewidth]{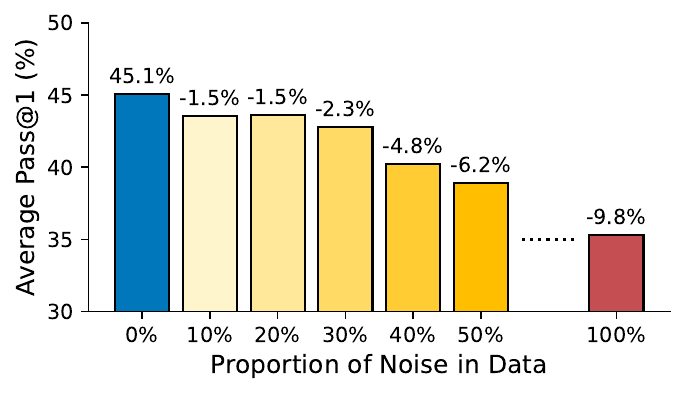}
        \caption{Noise results in increasing performance 
        degradation (1.5--9.8\%) as the noise proportion increases.}
        \label{fig:noise_rates}
    \end{minipage}
    \hfill
    \begin{minipage}{0.32\linewidth}
        \centering
        \includegraphics[width=\linewidth]{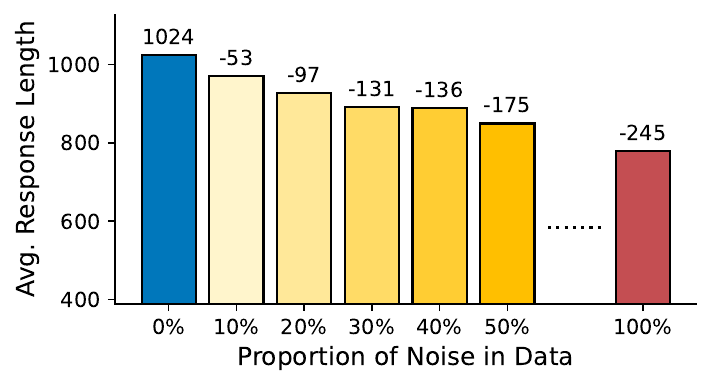}
        \caption{Training on noisy data results in shorter responses (5.2--23.9\%), indicating that noise induces weaker reasoning.}
        \label{fig:genlen}
    \end{minipage}
\end{figure*}

\minihead{Noisy data decreases pass@$k$.}
To determine if noisy data affects the range of problems a model can solve given 
multiple attempts, we evaluate the pass@$k$ of models trained on noisy data. We 
find that clean data expands the set of solvable problems, while noisy 
data shrinks it.  As shown in Figure \ref{fig:pass_k_noise}, the model trained 
on clean data consistently outperforms all other baselines by 4.6--16.6\% in 
terms of pass@$k$. However, the model trained on noisy data only outperforms the 
base model when $k=1$, while underperforming the base model by up to 6.1\% when 
$k>1$. This indicates that incorrect annotations reduce both sampling efficiency 
and the number of solvable problems, damaging the model's ``reasoning boundary'' 
\cite{yue2025does}.

\minihead{Noisy data increasingly leads to shorter reasoning.} Noise induces a 
collapse in reasoning length, leading to not only lower accuracy but also 
shorter responses. In Figure \ref{fig:noise_rates}, we show a monotonic drop in 
accuracy as noise increases, with degradation ranging from 1.5\% (at 10\% noise) 
to 9.8\% (at 100\% noise). This loss in accuracy is accompanied by a significant 
reduction in generation length. In Figure \ref{fig:genlen}, we show that models 
trained on noisy data produced responses that were 5.2\% to 23.9\% shorter than 
the model trained on clean data. This reduction implies that noisy data 
discourages complex reasoning chains, which weakens the reasoning capability.

\section{Algorithm Improvements Fail to Mitigate the Impact of Data Noise}
\label{sec:algs}

In this section, we investigate whether existing algorithm improvements mitigate 
the impact of noisy data on RLVR. We first introduce our experimental settings and 
then discuss our findings.

\subsection{Algorithm Selection}

We evaluated a set of state-of-the-art RLVR variants. We categorized these methods
into three algorithmic changes that hypothetically improves stability and robustness:
\vspace{-0.3em}
\begin{enumerate}[leftmargin=*,itemsep=1pt]
    \item \textbf{Gradient bias mitigation}: GRPO suffers from bias 
    caused by inherent policy loss \cite{liu2025understanding}, training-inference
    shifts \cite{yao2025offpolicy}, and noisy data \cite{cai2025reinforcement}. 
    Therefore, we include Dr.\ GRPO \cite{liu2025understanding}, truncated 
    importance sampling (TIS) \cite{yao2025offpolicy}, and policy gradient with 
    forward correction (PGFC) designed to address these bias, respectively.
    \item \textbf{Dynamic sampling}: GRPO suffers from vanishing 
    gradients when a rollout group has uniform rewards (all correct or all wrong). 
    To address this, prior work proposed dynamic sampling, which discard uniform 
    groups to focus on informative groups. Dynamic sampling may implicitly filter
    out failures caused by noise. Thus, we evaluate TIS \cite{yao2025offpolicy} 
    and DAPO \cite{yu2025dapo}, both of which incorporate dynamic sampling.
    \item \textbf{Clipping mechanisms}: Prior research suggests that asymmetric
    clipping can improve learning from noisy data by regulating entropy 
    \cite{lv2025climb}. Furthermore, adaptive clipping ratios based on group
    advantage can stabilize gradient updates when a batch of rollouts exhibit
    a variance of rewards. Therefore, we assess DAPO (asymmetric clipping) 
    \cite{yu2025dapo} and SAPO (adaptive clipping) \cite{gao2025soft}.
\end{enumerate}

In summary, we evaluate five algorithms to achieve a full coverage of the 
categories of state-of-the-art algorithmic improvements: Dr.\ GRPO (bias 
mitigation), TIS (bias mitigation + dynamic sampling), PGFC (bias mitigation), 
DAPO (dynamic sampling + clipping), and SAPO (clipping).

\begin{figure*}[t!]

    \centering
    \begin{subfigure}{0.33\textwidth}
        \includegraphics[width=\linewidth]{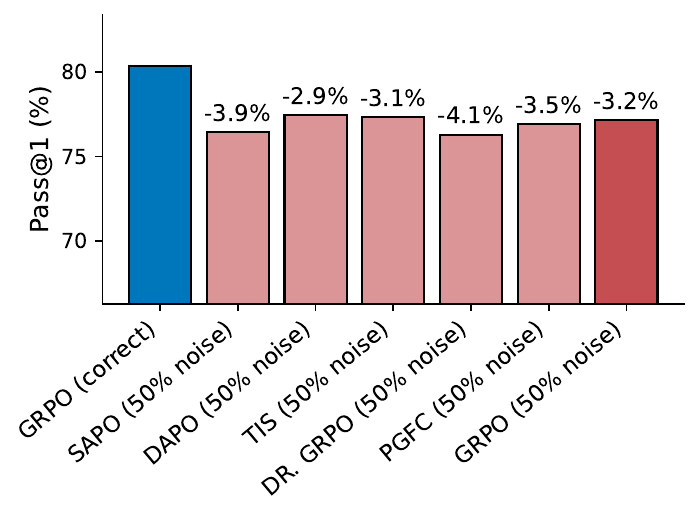}
        \caption{MATH-500.} 
        \label{fig:alg-math500}
    \end{subfigure}\hspace*{\fill}
    \begin{subfigure}{0.33\textwidth}
        \includegraphics[width=\linewidth]{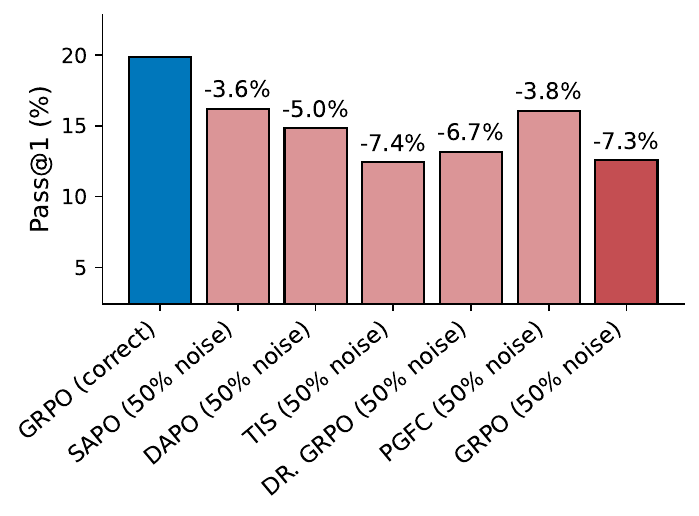}
        \caption{AIME 2024 and 2025.} 
        \label{fig:alg-aime}
    \end{subfigure}\hspace*{\fill}
    \begin{subfigure}{0.33\textwidth}
        \includegraphics[width=\linewidth]{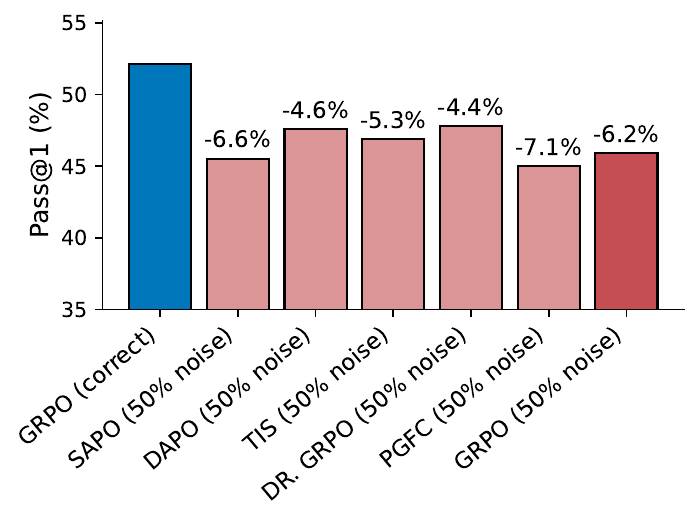}
        \caption{AMC 2023 and 2024.} 
        \label{fig:alg-amc}
    \end{subfigure}

    \caption{Existing algorithmic improvements fail to mitigate the impact of
    noisy data. Under 50\% noise, none of the improved algorithms achieved a 
    $>2$\% accuracy improvement on all benchmarks simultaneously compared to
    GRPO with 50\% noise, while all algorithms underperform the vanilla 
    GRPO with clean data by 3.1--7.3\%.} 
    \label{fig:algs}
\end{figure*}

\subsection{Experimental Settings}

In addition to the settings of Section \ref{sec:synthetic}, we use the default 
hyperparameter recommendations for
each algorithm based on their original publications. For TIS and DAPO, we used 
asymmetric clipping with an upper ratio of 0.28 and a lower ratio of 0.2, 
alongside an overlong buffer threshold of 2,048 tokens and a penalty factor of 
1.0. Additionally, for TIS, we applied an importance sampling cap of 2.0. For SAPO, 
we configured the adaptive clipping with a positive clipping parameter of 1.0 and 
a negative clipping parameter of 1.05. For PGFC, we used the ground-truth noise 
rates as the correction factors, measuring the method's upper-bound performance.

\subsection{Results}

\begin{myanswerbox}
    \textbf{\upshape Answer to RQ2}. Algorithmic improvements fail to compensate 
    for the noise in data. None of the SOTA variants (TIS, DAPO, SAPO, Dr.\ GRPO) 
    outperformed GRPO by more than 1\% under the 50\% noise. None of
    them consistently achieved performance within 5\% of GRPO on clean data. 
\end{myanswerbox}
\begin{wrapfigure}{r}{0.4\textwidth}
    \centering
    \includegraphics[width=\linewidth]{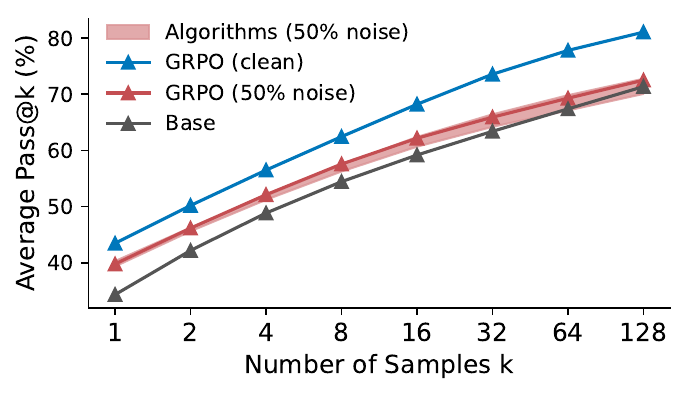}
    \caption{Evaluated algorithmic improvements fail to achieve a higher pass@$k$ than GRPO under 50\% noise.}
    \label{fig:alg_passk}
\end{wrapfigure}
\minihead{Algorithmic improvements fail to mitigate accuracy degradation.}
In Figure \ref{fig:algs}, we summarize the final accuracy achieved by each 
algorithm. Under 50\% noise, none of the evaluated algorithms achieved a 
consistent improvement over the GRPO (50\% noise) baseline. Specifically, 
no algorithm yielded more than a 2\% accuracy gain across all benchmarks 
simultaneously. Furthermore, all algorithmic variants underperform the model 
trained on clean data, with accuracy gaps ranging from 3.1\% to 7.3\%. This 
suggests that current algorithmic interventions are insufficient to compensate 
for the destructive effects of noise. We defer a detailed analysis of the
accuracy trend to Appendix \ref{sec:app-algs}, which shows similar trends to 
Figure \ref{fig:algs}.

\minihead{Algorithmic improvements fail to recover pass@$k$.}
To determine if algorithmic improvements could enable the models to solve a wider
range of problems given multiple attempts, we analyzed the average pass@$k$ 
performance of these methods across all benchmarks. In Figure \ref{fig:alg_passk}, 
we demonstrate that the performance of the improved algorithms (represented by 
the shaded red region) overlaps with the GRPO (50\% noise) baseline. 
None of the algorithmic improvements achieved a higher pass@$k$ score than the 
vanilla baseline, and all remained significantly below the GRPO with clean data. 
Additionally, we analyzed the response length of the trained models. We found 
that none of the models trained via improved algorithms consistently generates 
longer response than GRPO. These results indicate that 
improved algorithms fail to improve the model's reasoning boundary under noisy 
data. We defer more detailed results and analysis in Appendix 
\ref{sec:app-passk-algs} and \ref{sec:app-len-algs}.
\section{Impact of Real-world Annotation Noise}
\label{sec:real-world}
Moving beyond synthetic noisy data, we extend our analysis to the annotation 
errors found in real-world datasets. Specifically, we investigate the impact of 
naturally occurring annotation errors within the BIRD dataset \cite{li2023can} 
on RLVR performance.

\subsection{Experimental Settings}

\minihead{Training.}
We applied multi-turn RLVR with iterative query refinement \cite{liu2025skyrlsql}, and define a tri-level reward that yields 1 if the execution
results of the generated SQL and the gold SQL match, -1 if the output misses \textlangle solution\textrangle\ tags, and 0 otherwise. We trained each model 
for ten epochs (Appendix \ref{sec:app-ca}) with a batch size of 64, group size of 16, learning rate of $5\times 10^{-5}$, and LoRA rank of 32.

\minihead{Evaluation.}
Existing Text2SQL benchmarks suffer from substantial annotation errors 
\cite{liu2025nl2sql,pourreza2023evaluating,wretblad2024understanding,jin2026pervasive}. To ensure
our reported performance difference come from noisy training data rather than 
benchmark errors, we used the BIRD Mini-Dev set independently verified by Arcwise 
\cite{arcwise}. This benchmark contains corrections for 161 (32.3\%) of 
the original 498 instances. After independent, manual inspection by 
two of our authors, we find that the corrections by Arcwise are valid and sound.
We performed all evaluations using greedy decoding (temperature=0).

\subsection{Findings}

\begin{myanswerbox}
    \textbf{\upshape Answer to RQ3}. Real-world noisy data is destructive to 
    RLVR. Across different base models ranging from 
    32B to 685B parameters, training on noisy data results in up to 12\% lower
    accuracy compared to training on clean data.
\end{myanswerbox}
\begin{wrapfigure}{r}{0.44\textwidth}
    \centering
    \includegraphics[width=\linewidth]{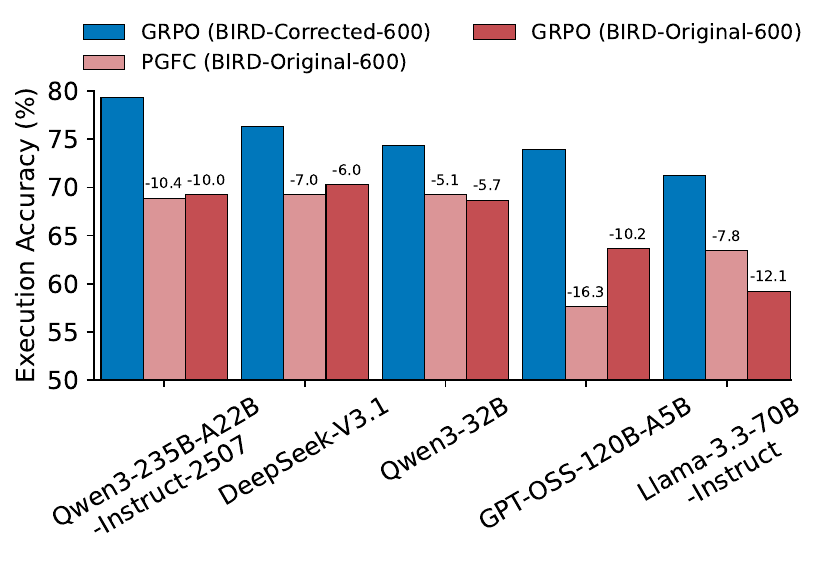}
    \caption{Real-world noise significantly degrades RLVR performance. Models 
    trained on the noisy BIRD dataset (BIRD-Original-600) via GRPO results in 
    5.7–12.1\% lower accuracy than BIRD-Corrected-600. PGFC 
    \cite{cai2025reinforcement}, an algorithm designed for noisy data, 
    fails to consistently improve over GRPO and remains 5.1–16.3\% behind the 
    clean data.}
    \label{fig:bird}
\end{wrapfigure}
\minihead{Real-world noise leads to significant degradation.}
As shown in Figure \ref{fig:bird}, we find that training on BIRD-Corrected-600 
demonstrates consistent and significant superiority over training on 
BIRD-Original-600. Models trained on the BIRD-Corrected-600 consistently 
outperform those trained on the original noisy data, with performance gaps 
ranging from a minimum of 5.7\% to a maximum of 12.1\%. This confirms that the 
impact of noisy data observed in synthetic settings transfers directly to 
realistic domains.

\minihead{Algorithmic mitigation fails to handle real-world noise.} 
We also tested whether PGFC \cite{cai2025reinforcement}, an algorithm 
specifically designed to mitigate impact of noise on RLVR, could recover the 
performance lost to real-world noise. Consistent with our findings from 
synthetic noise, we show that PGFC failed to yield consistent improvements over 
the GRPO. On GPT-OSS-120B-A5B, PGFC performed significantly worse 
than GRPO on noisy data. Overall, models trained with PGFC on 
BIRD-Original-600 remained 5.1-16.3\% less accurate than those trained on 
clean data.

\section{Conclusion}

In this work, we demonstrate the significant impact of noisy data on RLVR. 
Contrary to prior studies, we developed a rigorous data re-verification pipeline 
to construct a set of truly noisy data and investigated real-world noise
by correcting a subset of the noisy BIRD dataset. Empirically, we identify 
a consistent performance collapse of 8--10\% due to synthetic noise and 6--12\% 
due to real-world noise. We show that existing RLVR algorithms result in 
performance indistinguishable from the basic GRPO. These findings highlight 
that RLVR risks yielding suboptimal models without high-quality data, which 
has not yet been addressed by existing algorithmic interventions.

\section{Limitation and Impact Statement}
\label{sec:limit-impact}
By demonstrating that algorithmic interventions fail against noisy RLVR data, 
we highlight the critical necessity of high-quality data curation. To prevent 
deploying degraded models in high-stakes domains like science and code 
generation, researchers must prioritize data quality over algorithmic 
complexity. A primary limitation of our study is its focus on outcome-based 
rewards and deterministic tasks (math and Text2SQL). Future work could 
investigate whether this noise sensitivity persists in open-ended domains or 
when using process-based reward models that verify intermediate steps.

\begin{ack}
This research was 
supported in part by a sponsorship from Bridgewater AIA Labs.
\end{ack}

\bibliography{paper}

@misc{liu2025skyrlsql,
      title={SkyRL-SQL: Matching GPT-4o and o4-mini on Text2SQL with Multi-Turn RL},
      author={Shu Liu and Sumanth Hegde and Shiyi Cao and Alan Zhu and Dacheng Li and Tyler Griggs and Eric Tang and Akshay Malik and Kourosh Hakhamaneshi and Richard Liaw and Philipp Moritz and Matei Zaharia and Joseph E. Gonzalez and Ion Stoica},
      year={2025},
      note={Notion Blog}
}

@article{shao2024deepseekmath,
  title={Deepseekmath: Pushing the limits of mathematical reasoning in open language models},
  author={Shao, Zhihong and Wang, Peiyi and Zhu, Qihao and Xu, Runxin and Song, Junxiao and Bi, Xiao and Zhang, Haowei and Zhang, Mingchuan and Li, YK and Wu, Yang and others},
  journal={arXiv preprint arXiv:2402.03300},
  year={2024}
}

@article{shao2025spurious,
  title={Spurious rewards: Rethinking training signals in rlvr},
  author={Shao, Rulin and Li, Shuyue Stella and Xin, Rui and Geng, Scott and Wang, Yiping and Oh, Sewoong and Du, Simon Shaolei and Lambert, Nathan and Min, Sewon and Krishna, Ranjay and others},
  journal={arXiv preprint arXiv:2506.10947},
  year={2025}
}

@article{park2025clip,
  title={Clip-Low Increases Entropy and Clip-High Decreases Entropy in Reinforcement Learning of Large Language Models},
  author={Park, Jaesung R and Kim, Junsu and Kim, Gyeongman and Jo, Jinyoung and Choi, Sean and Cho, Jaewoong and Ryu, Ernest K},
  journal={arXiv preprint arXiv:2509.26114},
  year={2025}
}

@article{lv2025climb,
  title={The Climb Carves Wisdom Deeper Than the Summit: On the Noisy Rewards in Learning to Reason},
  author={Lv, Ang and Xie, Ruobing and Sun, Xingwu and Kang, Zhanhui and Yan, Rui},
  journal={arXiv preprint arXiv:2505.22653},
  year={2025}
}

@article{wen2025reinforcement,
  title={Reinforcement learning with verifiable rewards implicitly incentivizes correct reasoning in base llms},
  author={Wen, Xumeng and Liu, Zihan and Zheng, Shun and Ye, Shengyu and Wu, Zhirong and Wang, Yang and Xu, Zhijian and Liang, Xiao and Li, Junjie and Miao, Ziming and others},
  journal={arXiv preprint arXiv:2506.14245},
  year={2025}
}

@article{wei2025swe,
  title={Swe-rl: Advancing llm reasoning via reinforcement learning on open software evolution},
  author={Wei, Yuxiang and Duchenne, Olivier and Copet, Jade and Carbonneaux, Quentin and Zhang, Lingming and Fried, Daniel and Synnaeve, Gabriel and Singh, Rishabh and Wang, Sida I},
  journal={arXiv preprint arXiv:2502.18449},
  year={2025}
}

@misc{deepswe2025,
  title={DeepSWE: Training a State-of-the-Art Coding Agent from Scratch by Scaling RL},
  author={Michael Luo and Naman Jain and Jaskirat Singh and Sijun Tan and Ameen Patel and Qingyang Wu and Alpay Ariyak and Colin Cai and Tarun Venkat and Shang Zhu and Ben Athiwaratkun and Manan Roongta and Ce Zhang and Li Erran Li and Raluca Ada Popa and Koushik Sen and Ion Stoica},
  howpublished={\url{https://pretty-radio-b75.notion.site/DeepSWE-Training-a-Fully-Open-sourced-State-of-the-Art-Coding-Agent-by-Scaling-RL-22281902c1468193aabbe9a8c59bbe33}},
  note={Notion Blog},
  year={2025}
}

@misc{deepscaler2025,
  title={DeepScaleR: Surpassing O1-Preview with a 1.5B Model by Scaling RL},
  author={Michael Luo and Sijun Tan and Justin Wong and Xiaoxiang Shi and William Y. Tang and Manan Roongta and Colin Cai and Jeffrey Luo and Li Erran Li and Raluca Ada Popa and Ion Stoica},
  howpublished={\url{https://pretty-radio-b75.notion.site/DeepScaleR-Surpassing-O1-Preview-with-a-1-5B-Model-by-Scaling-RL-19681902c1468005bed8ca303013a4e2}},
  note={Notion Blog},
  year={2025}
}

@article{ma2025general,
  title={General-reasoner: Advancing llm reasoning across all domains},
  author={Ma, Xueguang and Liu, Qian and Jiang, Dongfu and Zhang, Ge and Ma, Zejun and Chen, Wenhu},
  journal={arXiv preprint arXiv:2505.14652},
  year={2025}
}

@misc{deepcoder2025,
  title={DeepCoder: A Fully Open-Source 14B Coder at O3-mini Level},
  author={Michael Luo and Sijun Tan and Roy Huang and Ameen Patel and Alpay Ariyak and Qingyang Wu and Xiaoxiang Shi and Rachel Xin and Colin Cai and Maurice Weber and Ce Zhang and Li Erran Li and Raluca Ada Popa and Ion Stoica},
  howpublished={\url{https://pretty-radio-b75.notion.site/DeepCoder-A-Fully-Open-Source-14B-Coder-at-O3-mini-Level-1cf81902c14680b3bee5eb349a512a51}},
  note={Notion Blog},
  year={2025}
}

@article{su2025crossing,
  title={Crossing the Reward Bridge: Expanding RL with Verifiable Rewards Across Diverse Domains},
  author={Su, Yi and Yu, Dian and Song, Linfeng and Li, Juntao and Mi, Haitao and Tu, Zhaopeng and Zhang, Min and Yu, Dong},
  journal={arXiv preprint arXiv:2503.23829},
  year={2025}
}

@article{guo2025deepseek,
  title={Deepseek-r1: Incentivizing reasoning capability in llms via reinforcement learning},
  author={Guo, Daya and Yang, Dejian and Zhang, Haowei and Song, Junxiao and Zhang, Ruoyu and Xu, Runxin and Zhu, Qihao and Ma, Shirong and Wang, Peiyi and Bi, Xiao and others},
  journal={arXiv preprint arXiv:2501.12948},
  year={2025}
}

@online{openaio3,
  title={Introducing OpenAI o3 and o4-mini},
  author={OpenAI},
  year={2025},
  url={https://openai.com/index/introducing-o3-and-o4-mini/},
  note={Accessed Dec. 10, 2025}}

@article{li2023can,
  title={Can llm already serve as a database interface? a big bench for large-scale database grounded text-to-sqls},
  author={Li, Jinyang and Hui, Binyuan and Qu, Ge and Yang, Jiaxi and Li, Binhua and Li, Bowen and Wang, Bailin and Qin, Bowen and Geng, Ruiying and Huo, Nan and others},
  journal={Advances in Neural Information Processing Systems},
  volume={36},
  pages={42330--42357},
  year={2023}
}

@article{liu2025understanding,
  title={Understanding r1-zero-like training: A critical perspective},
  author={Liu, Zichen and Chen, Changyu and Li, Wenjun and Qi, Penghui and Pang, Tianyu and Du, Chao and Lee, Wee Sun and Lin, Min},
  journal={arXiv preprint arXiv:2503.20783},
  year={2025}
}

@article{yu2025dapo,
  title={Dapo: An open-source llm reinforcement learning system at scale},
  author={Yu, Qiying and Zhang, Zheng and Zhu, Ruofei and Yuan, Yufeng and Zuo, Xiaochen and Yue, Yu and Dai, Weinan and Fan, Tiantian and Liu, Gaohong and Liu, Lingjun and others},
  journal={arXiv preprint arXiv:2503.14476},
  year={2025}
}

@article{gao2025soft,
  title={Soft Adaptive Policy Optimization},
  author={Gao, Chang and Zheng, Chujie and Chen, Xiong-Hui and Dang, Kai and Liu, Shixuan and Yu, Bowen and Yang, An and Bai, Shuai and Zhou, Jingren and Lin, Junyang},
  journal={arXiv preprint arXiv:2511.20347},
  year={2025}
}

@article{chen2025minimax,
  title={MiniMax-M1: Scaling Test-Time Compute Efficiently with Lightning Attention},
  author={Chen, Aili and Li, Aonian and Gong, Bangwei and Jiang, Binyang and Fei, Bo and Yang, Bo and Shan, Boji and Yu, Changqing and Wang, Chao and Zhu, Cheng and others},
  journal={arXiv preprint arXiv:2506.13585},
  year={2025}
}

@article{team2025kimi,
  title={Kimi k2: Open agentic intelligence},
  author={Team, Kimi and Bai, Yifan and Bao, Yiping and Chen, Guanduo and Chen, Jiahao and Chen, Ningxin and Chen, Ruijue and Chen, Yanru and Chen, Yuankun and Chen, Yutian and others},
  journal={arXiv preprint arXiv:2507.20534},
  year={2025}
}

@misc{yao2025offpolicy,
  title = {Your Efficient RL Framework Secretly Brings You Off-Policy RL Training},
  url = {https://fengyao.notion.site/off-policy-rl},
  author = {Yao, Feng and Liu, Liyuan and Zhang, Dinghuai and Dong, Chengyu and Shang, Jingbo and Gao, Jianfeng},
  journal = {Feng Yao's Notion},
  year = {2025},
  month = aug,
}

@misc{griggs2025skrylv01,
      title={Evolving SkyRL into a Highly-Modular RL Framework},
      author={Tyler Griggs and Sumanth Hegde and Eric Tang and Shu Liu and Shiyi Cao and Dacheng Li and Charlie Ruan and Philipp Moritz and Kourosh Hakhamaneshi and Richard Liaw and Akshay Malik and Matei Zaharia and Joseph E. Gonzalez and Ion Stoica},
      year={2025},
      note={Notion Blog}
}

@article{cai2025reinforcement,
  title={Reinforcement learning with verifiable yet noisy rewards under imperfect verifiers},
  author={Cai, Xin-Qiang and Wang, Wei and Liu, Feng and Liu, Tongliang and Niu, Gang and Sugiyama, Masashi},
  journal={arXiv preprint arXiv:2510.00915},
  year={2025}
}

@article{mansouri2025noise,
  title={Noise-corrected GRPO: From Noisy Rewards to Unbiased Gradients},
  author={Mansouri, Omar El and Seddik, Mohamed El Amine and Lahlou, Salem},
  journal={arXiv preprint arXiv:2510.18924},
  year={2025}
}

@inproceedings{wretblad2024understanding,
  title={Understanding the Effects of Noise in Text-to-SQL: An Examination of the BIRD-Bench Benchmark},
  author={Wretblad, Niklas and Riseby, Fredrik and Biswas, Rahul and Ahmadi, Amin and Holmstr{\"o}m, Oskar},
  booktitle={Proceedings of the 62nd Annual Meeting of the Association for Computational Linguistics (Volume 2: Short Papers)},
  pages={356--369},
  year={2024}
}

@inproceedings{liu2025nl2sql,
  title={Nl2sql-bugs: A benchmark for detecting semantic errors in nl2sql translation},
  author={Liu, Xinyu and Shen, Shuyu and Li, Boyan and Tang, Nan and Luo, Yuyu},
  booktitle={Proceedings of the 31st ACM SIGKDD Conference on Knowledge Discovery and Data Mining V. 2},
  pages={5662--5673},
  year={2025}
}

@inproceedings{pourreza2023evaluating,
  title={Evaluating Cross-Domain Text-to-SQL Models and Benchmarks},
  author={Pourreza, Mohammadreza and Rafiei, Davood},
  booktitle={Proceedings of the 2023 Conference on Empirical Methods in Natural Language Processing},
  pages={1601--1611},
  year={2023}
}

@article{beyer2020we,
  title={Are we done with imagenet?},
  author={Beyer, Lucas and H{\'e}naff, Olivier J and Kolesnikov, Alexander and Zhai, Xiaohua and Oord, A{\"a}ron van den},
  journal={arXiv preprint arXiv:2006.07159},
  year={2020}
}

@article{yu2025utboost,
  title={Utboost: Rigorous evaluation of coding agents on swe-bench},
  author={Yu, Boxi and Zhu, Yuxuan and He, Pinjia and Kang, Daniel},
  journal={ACL 2025},
  year={2025}
}

@online{mathVerify,
  title={Math Verify},
  author={Kydlíček, Hynek and Gandenberger, Greg and Maksin, Leon},
  year={2025},
  url={https://github.com/huggingface/Math-Verify},
  note={Accessed Dec. 10, 2025}
}

@misc{tinker,
  title={Tinker},
  author={Thinking Machines Team},
  url={https://thinkingmachines.ai/tinker/},
  year={2025},
  note={Accessed Dec. 10, 2025}
}

@article{schulman2025lora,
  author = {John Schulman and Thinking Machines Lab},
  title = {LoRA Without Regret},
  journal = {Thinking Machines Lab: Connectionism},
  year = {2025},
  note = {https://thinkingmachines.ai/blog/lora/},
  doi = {10.64434/tml.20250929},
}

@inproceedings{lightman2023let,
  title={Let's verify step by step},
  author={Lightman, Hunter and Kosaraju, Vineet and Burda, Yuri and Edwards, Harrison and Baker, Bowen and Lee, Teddy and Leike, Jan and Schulman, John and Sutskever, Ilya and Cobbe, Karl},
  booktitle={ICLR},
  year={2023}
}

@misc{aime,
  title={AIME Problems and Solutions},
  author={AoPS Online Community},
  url={https://artofproblemsolving.com/wiki/index.php/AIME_Problems_and_Solutions},
  year={2025},
  note={Accessed Dec. 10, 2025}
}

@misc{amc,
  title={AMC Problems and Solutions},
  author={AoPS Online Community},
  url={https://artofproblemsolving.com/wiki/index.php?title=AMC_Problems_and_Solutions},
  year={2025},
  note={Accessed Dec. 10, 2025}
}

@misc{arcwise,
  author = {Arcwise},
  title = {BIRD minidev - corrections},
  year = {2025},
  url = {https://docs.google.com/spreadsheets/d/1IGm9Otruey60ujUnl8AOkepY3qgWHdFJHnX7hQGUeCw},
  note = {Accessed: 2025-09-15}
}

@article{natarajan2013learning,
  title={Learning with noisy labels},
  author={Natarajan, Nagarajan and Dhillon, Inderjit S and Ravikumar, Pradeep K and Tewari, Ambuj},
  journal={Advances in neural information processing systems},
  volume={26},
  year={2013}
}

@article{zhang2018generalized,
  title={Generalized cross entropy loss for training deep neural networks with noisy labels},
  author={Zhang, Zhilu and Sabuncu, Mert},
  journal={Advances in neural information processing systems},
  volume={31},
  year={2018}
}

@article{newcombe1998two,
  title={Two-sided confidence intervals for the single proportion: comparison of seven methods},
  author={Newcombe, Robert G},
  journal={Statistics in medicine},
  volume={17},
  number={8},
  pages={857--872},
  year={1998},
  publisher={Wiley Online Library}
}

@article{yue2025does,
  title={Does reinforcement learning really incentivize reasoning capacity in llms beyond the base model?},
  author={Yue, Yang and Chen, Zhiqi and Lu, Rui and Zhao, Andrew and Wang, Zhaokai and Song, Shiji and Huang, Gao},
  journal={arXiv preprint arXiv:2504.13837},
  year={2025}
}

@article{jin2026pervasive,
  title={Pervasive Annotation Errors Break Text-to-SQL Benchmarks and Leaderboards},
  author={Jin, Tengjun and Choi, Yoojin and Zhu, Yuxuan and Kang, Daniel},
  journal={VLDB 2026},
  year={2026}
}
\bibliographystyle{icml2025}

\newpage
\appendix
\onecolumn
\section{Detailed Experimental Results}

\subsection{Training Dynamics}
\label{sec:app-dyna}

\begin{figure*}[t!]
    \centering
    \begin{subfigure}{0.45\textwidth}
        \includegraphics[width=\linewidth]{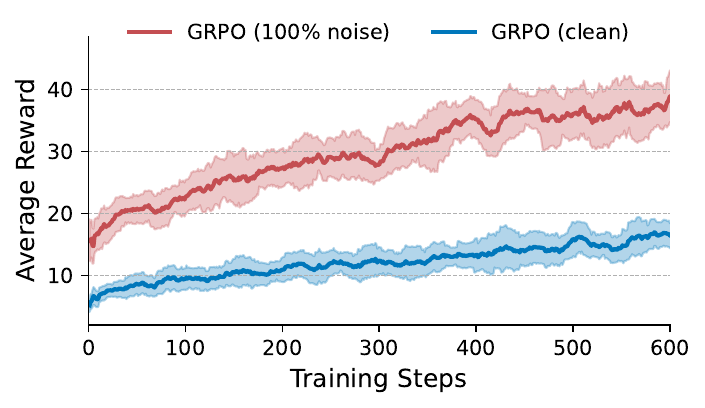}
        \caption{Average reward.} 
        \label{fig:dynamics-reward}
    \end{subfigure}
    \begin{subfigure}{0.45\textwidth}
        \includegraphics[width=\linewidth]{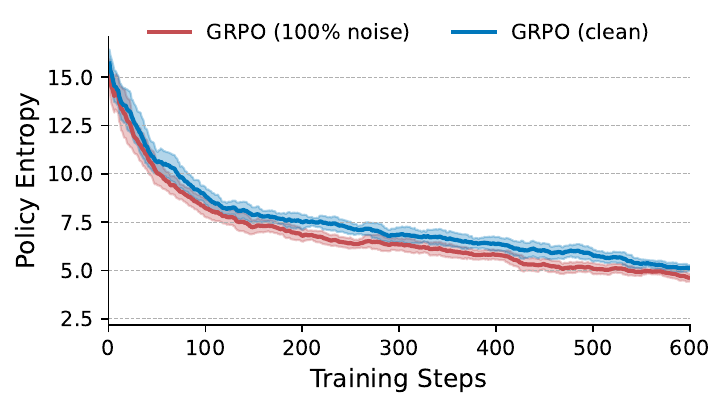}
        \caption{Policy entropy.} 
        \label{fig:dynamics-entropy}
    \end{subfigure}

    \caption{Training dynamics of RLVR under clean versus 100\% noisy data. (a) The model achieves a 149\% higher average reward when trained on noisy data, indicating that it easily exploits and satisfies its own flawed priors. (b) Training on noise induces a more rapid decline in policy entropy (9\% lower on average), reflecting premature convergence and diminished exploration.} 
    \label{fig:app-dynamics}
\end{figure*}

To understand the mechanism behind the severe performance degradation caused by 
noisy data, we analyze the training dynamics of the model. Specifically, we 
compare the average reward and policy entropy curves of the GRPO algorithm
on 100\% verified noise versus clean data (Figure~\ref{fig:app-dynamics}).

\minihead{Reward Exploitation on Flawed Priors.} The model finds it 
significantly easier to optimize for incorrect annotations than for correct ones. 
As shown in Figure~\ref{fig:dynamics-reward}, during training, the model 
achieves 149\% higher average rewards on noisy data compared to clean data. 
Because the synthetic noise in the dataset originates from the base model's own 
incorrect generations, these flawed reasoning paths already possess a high 
initial probability under the model's pre-trained distribution. Consequently, 
the model effortlessly rediscovers and satisfies these incorrect labels, 
exploiting its own flawed priors rather than learning novel, rigorous reasoning 
chains.

\minihead{Premature Entropy Collapse.} This ease of optimization actively harms 
the exploration necessary for RLVR. In Figure \ref{fig:dynamics-entropy},
we illustrate the evolution of policy entropy, which serves as a proxy for the 
model's exploratory behavior. We observe that training on noise results in a 
consistently lower policy entropy by 9\% on average, compared to training on 
clean data. The rapid collapse in entropy indicates that the model becomes 
overconfident in its generated outputs much earlier in the training process, 
prematurely curtailing its search space.

\minihead{The Confirmation Bias Loop.} Together, these dynamics reveal that 
applying RLVR to model-generated noise creates a pathological ``confirmation 
bias loop.'' Because the model is heavily and easily rewarded for reasoning 
paths it already favors, it rapidly loses the incentive to explore alternative, 
correct solutions. Instead of expanding the model's reasoning capabilities, RLVR 
with noisy data merely reinforces and entrenches the model's pre-existing errors, 
leading to the collapsed reasoning lengths and degraded accuracy observed in our 
evaluations.

\subsection{Pass@k Results of Models Trained on Noisy Data}
\label{sec:app-passk}

\begin{figure*}[t!]
    \centering
    \begin{subfigure}{0.33\textwidth}
        \includegraphics[width=\linewidth]{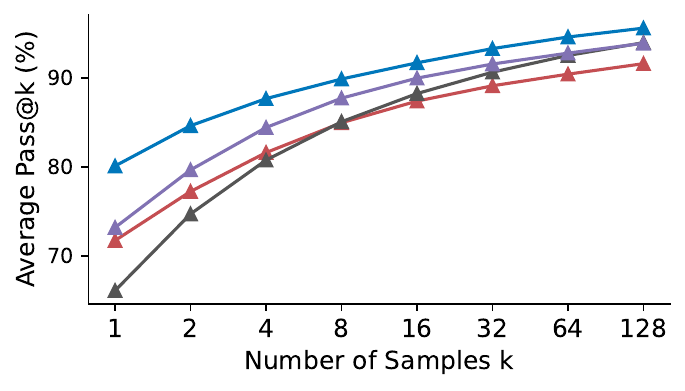}
        \caption{MATH-500.} 
        \label{fig:math500-passk}
    \end{subfigure}\hspace*{\fill}
    \begin{subfigure}{0.33\textwidth}
        \includegraphics[width=\linewidth]{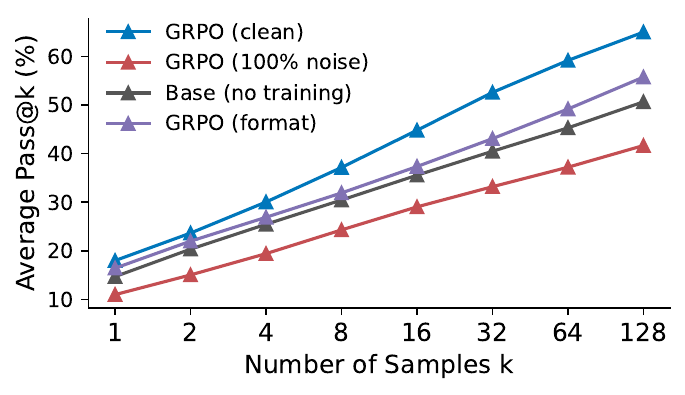}
        \caption{AIME 2024 and 2025.} 
        \label{fig:aime-passk}
    \end{subfigure}\hspace*{\fill}
    \begin{subfigure}{0.33\textwidth}
        \includegraphics[width=\linewidth]{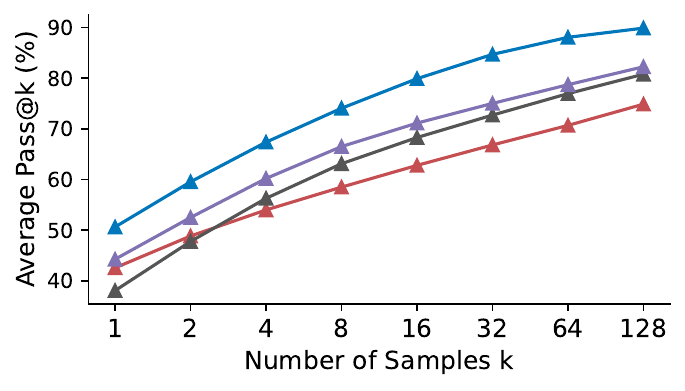}
        \caption{AMC 2023 and 2024.} 
        \label{fig:amc-passk}
    \end{subfigure}

    \caption{Noise leads to lower pass@$k$ than the base model when $k > 1$, showing that
noise does not improve capability boundary.} 
    \label{fig:app-passk}
\end{figure*}

We show the pass@$k$ performance of models trained on noisy data in Figure \ref{fig:app-passk}.
While the model trained on correct annotations (blue) consistently expands the 
solution space across all $k$, the model trained on 100\% noise (red) exhibits 
failures consistently. On easier benchmarks like MATH-500 (Figure 
\ref{fig:math500-passk}), the noisy model performs comparably to the base 
model at $k=1$, but falls behind as $k$ increases. This trend is even more 
severe on complex reasoning benchmarks like AIME (Figure \ref{fig:aime-passk}) 
and AMC (Figure \ref{fig:amc-passk}), where the base model (grey) 
consistently outperforms the noisy model across nearly all values of $k$. 

This 
crossover indicates that while noise might improve greedy decoding ($k=1$) 
slightly, it degrades the model's reasoning capabilities in general. Furthermore, 
training with format-only rewards (purple) consistently outperforms training with 
verifiable noise, confirming that the ``signal'' in noisy data is negative, 
dragging performance below what could be achieved by simply teaching the model 
the correct output format.

\subsection{Results of Training on Different Noise Rates}
\label{sec:app-noise-rates}

\begin{figure*}[t!]
    \centering
    \begin{subfigure}{0.33\textwidth}
        \includegraphics[width=\linewidth]{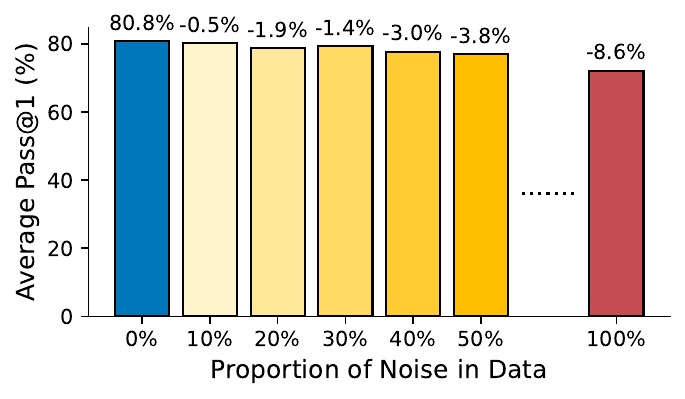}
        \caption{MATH-500.} 
        \label{fig:noise-math500}
    \end{subfigure}\hspace*{\fill}
    \begin{subfigure}{0.33\textwidth}
        \includegraphics[width=\linewidth]{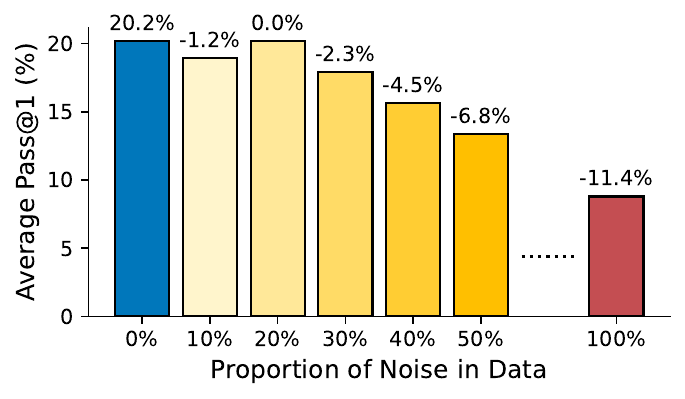}
        \caption{AIME 2024 and 2025.} 
        \label{fig:noise-aime}
    \end{subfigure}\hspace*{\fill}
    \begin{subfigure}{0.33\textwidth}
        \includegraphics[width=\linewidth]{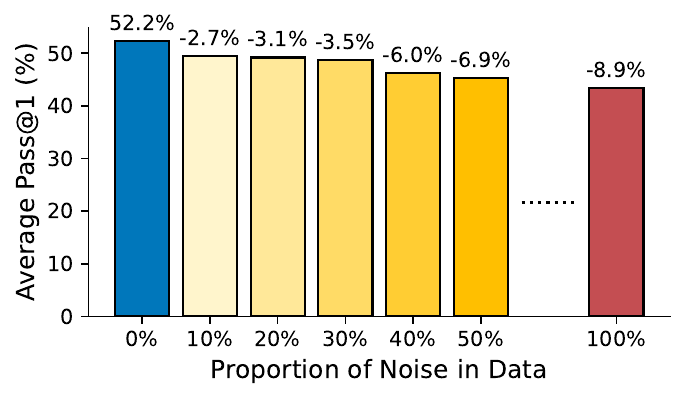}
        \caption{AMC 2023 and 2024.} 
        \label{fig:noise-amc}
    \end{subfigure}

    \caption{Training on noisy data results in increasing performance 
        degradation (by up to 11.4\%) as the noise proportion increases.} 
    \label{fig:app-noise-rates}
\end{figure*}

\begin{figure*}[t!]
    \centering
    \begin{subfigure}{0.33\textwidth}
        \includegraphics[width=\linewidth]{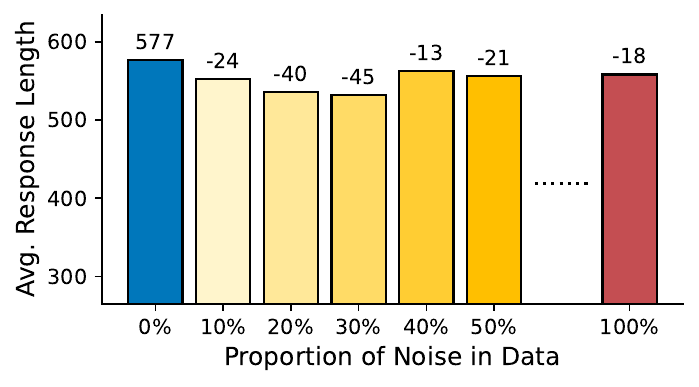}
        \caption{MATH-500.} 
        \label{fig:length-math500}
    \end{subfigure}\hspace*{\fill}
    \begin{subfigure}{0.33\textwidth}
        \includegraphics[width=\linewidth]{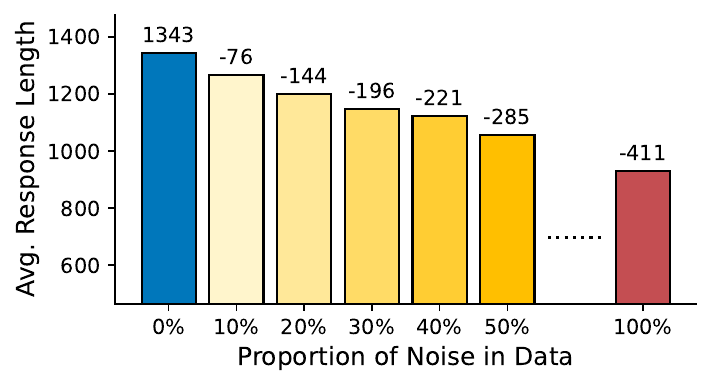}
        \caption{AIME 2024 and 2025.} 
        \label{fig:length-aime}
    \end{subfigure}\hspace*{\fill}
    \begin{subfigure}{0.33\textwidth}
        \includegraphics[width=\linewidth]{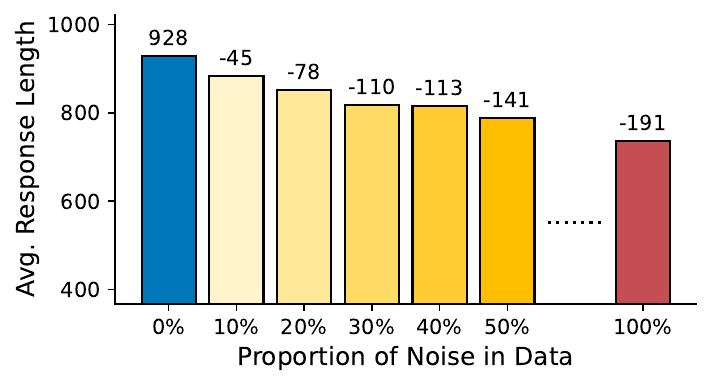}
        \caption{AMC 2023 and 2024.} 
        \label{fig:length-amc}
    \end{subfigure}

    \caption{Training on noisy data results in shorter responses, indicating that noise induces weaker reasoning.} 
    \label{fig:app-length}
\end{figure*}

\begin{figure*}[t!]
    \begin{subfigure}{0.32\textwidth}
        \includegraphics[width=\linewidth]{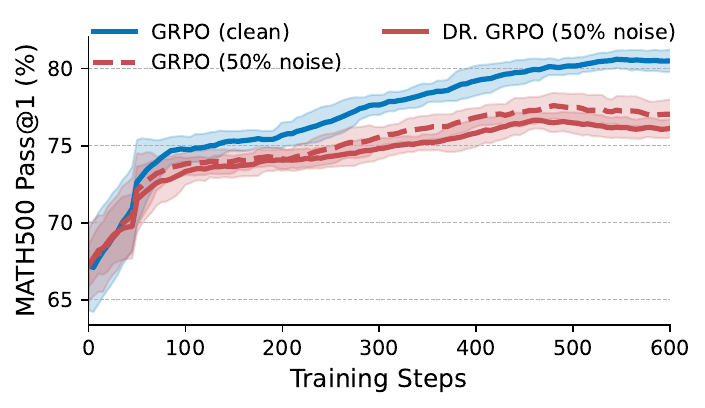}
    \end{subfigure}\hspace*{\fill}
    \begin{subfigure}{0.32\textwidth}
        \includegraphics[width=\linewidth]{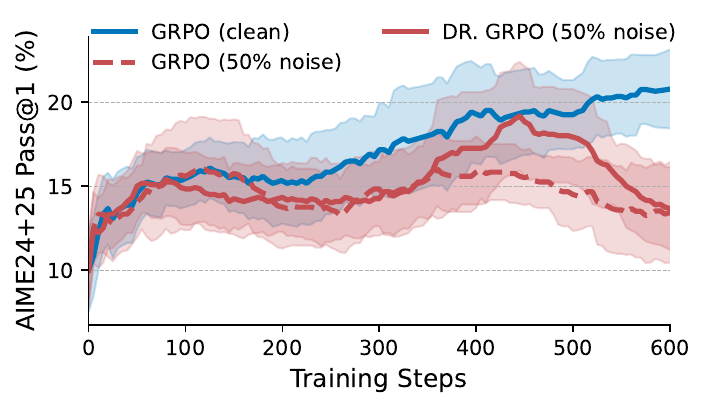}
    \end{subfigure}\hspace*{\fill}
    \begin{subfigure}{0.32\textwidth}
        \includegraphics[width=\linewidth]{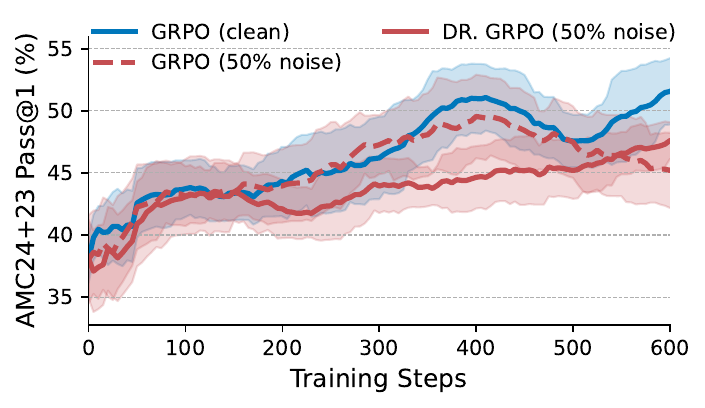}
    \end{subfigure}

    \begin{subfigure}{0.32\textwidth}
        \includegraphics[width=\linewidth]{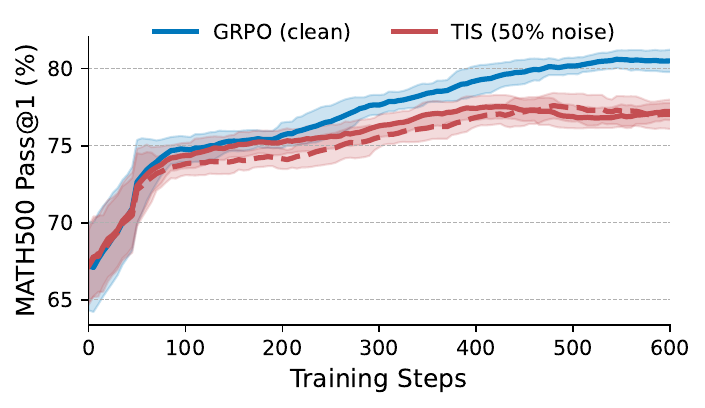}
    \end{subfigure}\hspace*{\fill}
    \begin{subfigure}{0.32\textwidth}
        \includegraphics[width=\linewidth]{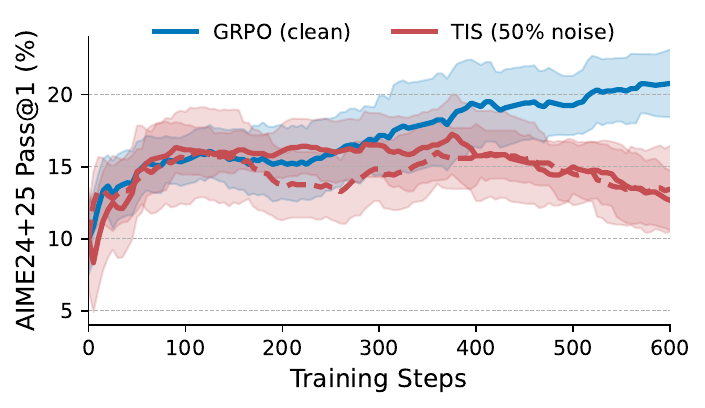}
    \end{subfigure}\hspace*{\fill}
    \begin{subfigure}{0.32\textwidth}
        \includegraphics[width=\linewidth]{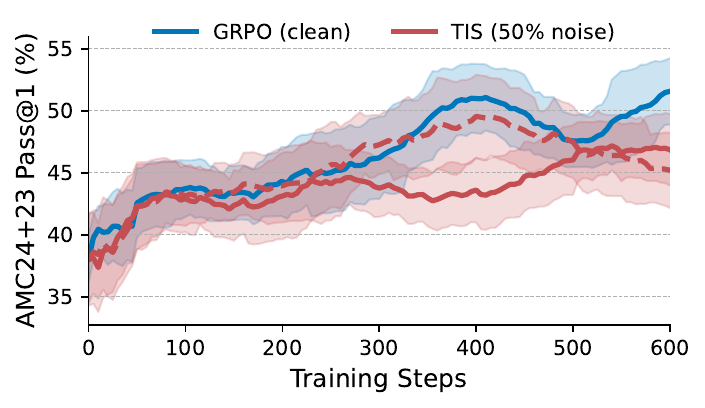}
    \end{subfigure}

    \begin{subfigure}{0.32\textwidth}
        \includegraphics[width=\linewidth]{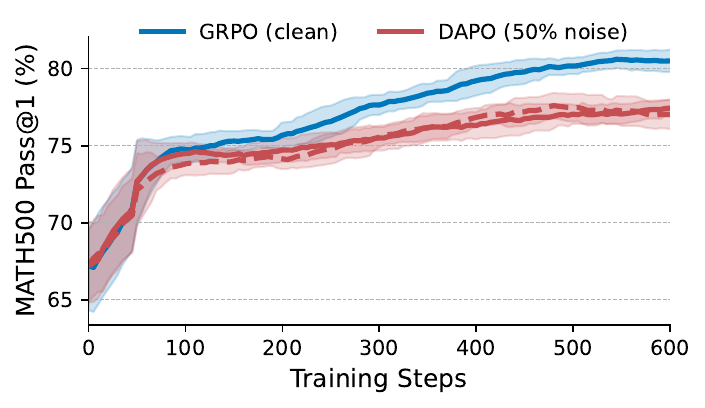}
    \end{subfigure}\hspace*{\fill}
    \begin{subfigure}{0.32\textwidth}
        \includegraphics[width=\linewidth]{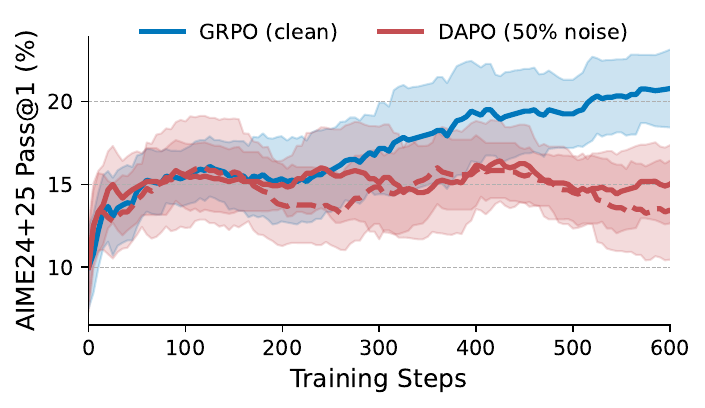}
    \end{subfigure}\hspace*{\fill}
    \begin{subfigure}{0.32\textwidth}
        \includegraphics[width=\linewidth]{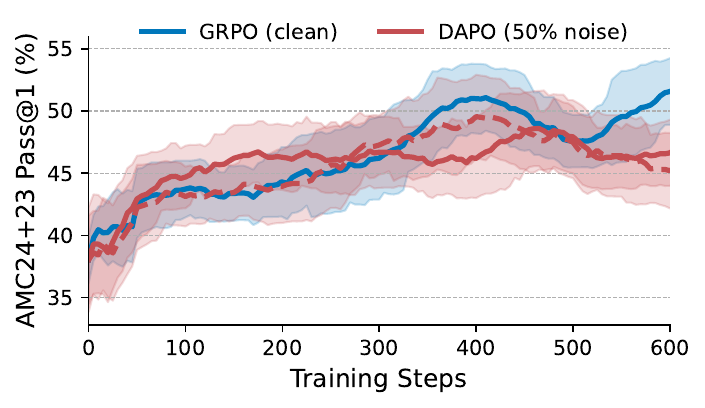}
    \end{subfigure}

    \begin{subfigure}{0.32\textwidth}
        \includegraphics[width=\linewidth]{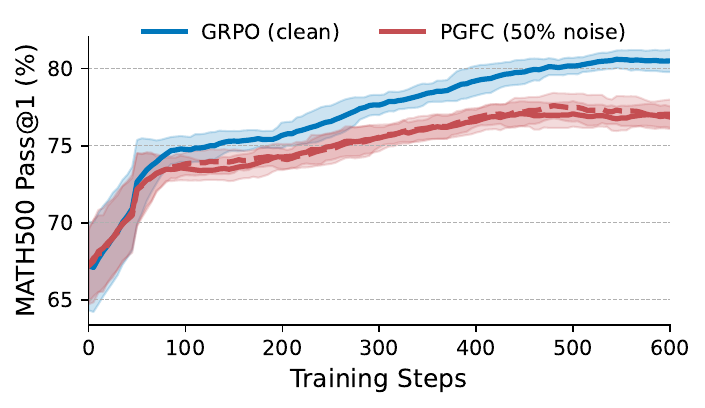}
    \end{subfigure}\hspace*{\fill}
    \begin{subfigure}{0.32\textwidth}
        \includegraphics[width=\linewidth]{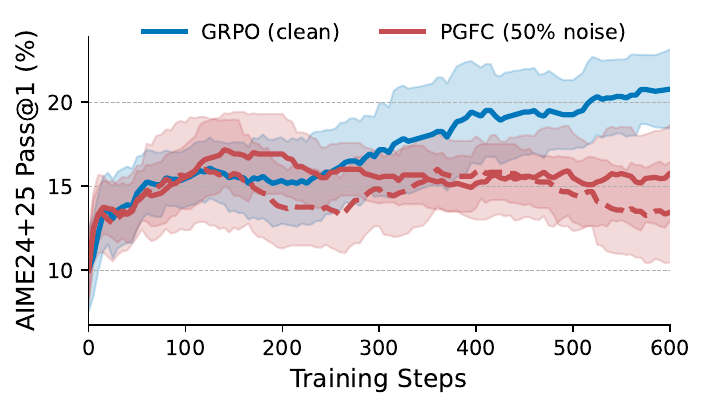}
    \end{subfigure}\hspace*{\fill}
    \begin{subfigure}{0.32\textwidth}
        \includegraphics[width=\linewidth]{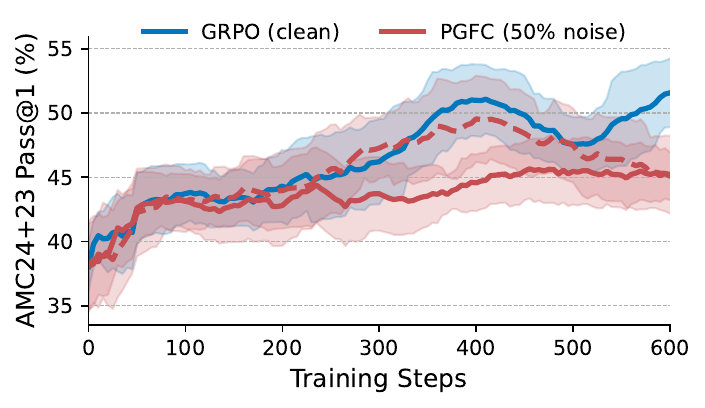}
    \end{subfigure}

    \begin{subfigure}{0.32\textwidth}
        \includegraphics[width=\linewidth]{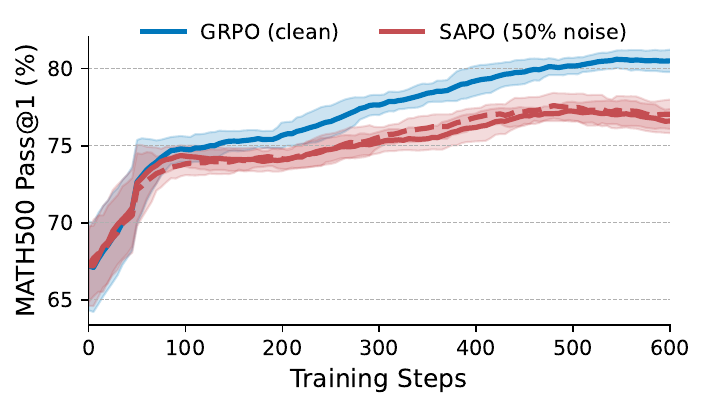}
        \caption{MATH-500.} 
        \label{fig:app-alg-math500}
    \end{subfigure}\hspace*{\fill}
    \begin{subfigure}{0.32\textwidth}
        \includegraphics[width=\linewidth]{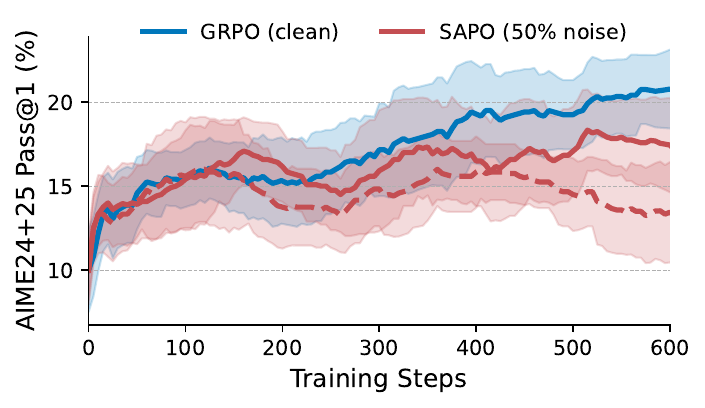}
        \caption{AIME 2024 and 2025.} 
        \label{fig:app-alg-aime}
    \end{subfigure}\hspace*{\fill}
    \begin{subfigure}{0.32\textwidth}
        \includegraphics[width=\linewidth]{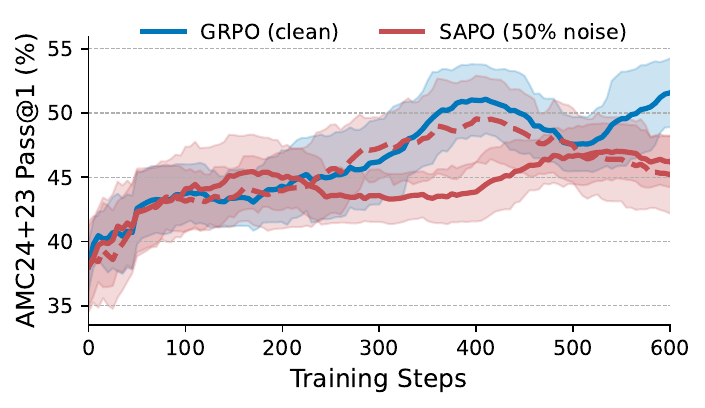}
        \caption{AMC 2023 and 2024.} 
        \label{fig:app-alg-amc}
    \end{subfigure}
\caption{Existing algorithm improvements fail to mitigate the impact of
    noisy data. Under 50\% noise, none of the improved algorithm achieved 
    $>2$\% accuracy improvement on all benchmarks simultaneously compared to
    GRPO with 50\% noise, while all algorithms underperform the vanilla 
    GRPO with clean data by 3.1-7.3\%} 
    \label{fig:app-algs}
\end{figure*}

We show the performance of models trained on data with different noise rates in 
Figure \ref{fig:app-noise-rates}.
As shown, we find a strict negative correlation: as the proportion of incorrect 
annotations rises, accuracy consistently declines. This degradation is most 
severe on the challenging AIME benchmark (Figure \ref{fig:noise-aime}), where 
100\% noise precipitates an 11.4\% collapse in accuracy compared to the clean 
baseline. Similarly, MATH-500 and AMC suffer significant drops of 8.6\% and 8.9\%, 
respectively. Crucially, we observe no ``safe'' noise threshold; even moderate
noise levels (e.g., 20–30\%) result in measurable performance penalties, 
confirming that RLVR lacks intrinsic robustness to label noise.

We show the average generation length of models trained on data with different 
noise rates in Figure \ref{fig:app-length}. We reveal a clear inverse 
relationship: as noise increases, reasoning chains become progressively shorter.
This effect is most profound on the complex AIME benchmark (Figure 
\ref{fig:length-aime}), where high-quality reasoning typically requires 
extensive steps. Here, we find a steep decline from 1343 tokens (clean data) to 
just 932 tokens (100\% noise), a loss of over 400 tokens or $\approx$30\% of the 
reasoning length. Even minor contamination (10\% noise) causes a noticeable drop 
of 76 tokens. This suggests that noisy rewards actively discourage the model from 
performing the deep exploration required for hard problems, instead biasing it 
toward shorter and likely incorrect solutions.

\subsection{Results of Different RLVR Algorithms}
\label{sec:app-algs}

In Figure \ref{fig:app-algs}, we show the test accuracy trajectories (pass@1) 
over all training steps for all algorithms. We first show a sharp contrast 
between clean and noisy supervision. Across all five algorithmic variants 
(rows) and three benchmarks (columns), the model trained on clean data 
establishes a clear performance upper bound, maintaining a steady upward 
trajectory. In contrast, the advanced algorithms (solid red) fail to differ 
from the GRPO baseline trained on the same noisy data (dashed red). The 
trajectories are nearly indistinguishable, with overlapping confidence intervals 
throughout the process.

Specifically, on the challenging AIME benchmark (middle column), both the 
advanced algorithms and the vanilla noisy baseline exhibit signs of training 
collapse after 400 steps, whereas the clean model continues to improve. This 
confirms that current algorithmic state-of-the-art methods, including Dr.\ GRPO, 
TIS, DAPO, PGFC, and SAPO, offer no significant robustness against validated 
label noise.

\subsection{Pass@k Results of Different RLVR Algorithms}
\label{sec:app-passk-algs}

\begin{figure*}[t!]
    \centering
    \begin{subfigure}{0.33\textwidth}
        \includegraphics[width=\linewidth]{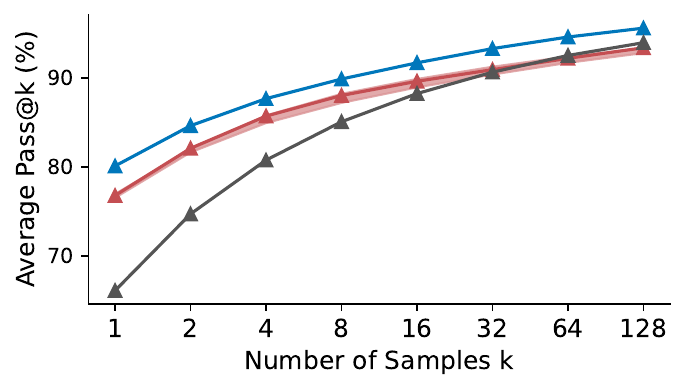}
        \caption{MATH-500.} 
        \label{fig:alg-passk-math500}
    \end{subfigure}\hspace*{\fill}
    \begin{subfigure}{0.33\textwidth}
        \includegraphics[width=\linewidth]{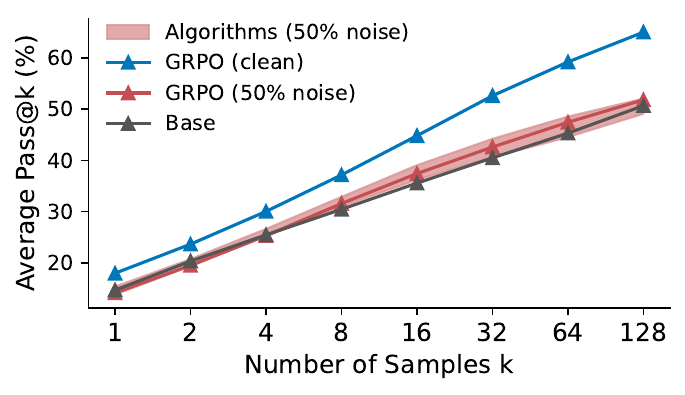}
        \caption{AIME 2024 and 2025.} 
        \label{fig:alg-passk-aime}
    \end{subfigure}\hspace*{\fill}
    \begin{subfigure}{0.33\textwidth}
        \includegraphics[width=\linewidth]{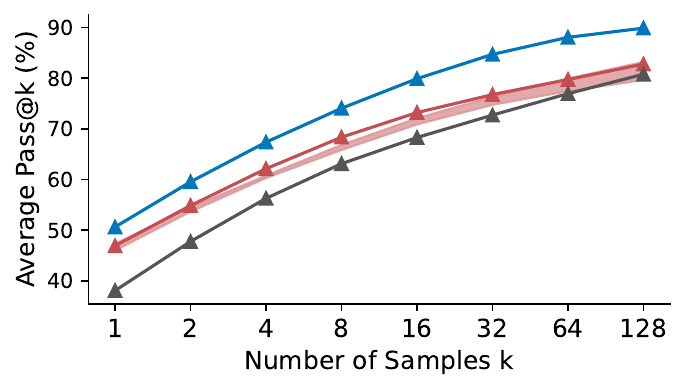}
        \caption{AMC 2023 and 2024.} 
        \label{fig:alg-passk-amc}
    \end{subfigure}

    \caption{None of the evaluated algorithmic improvements achieve higher 
    pass@$k$ score than GRPO under 50\% noise.} 
    \label{fig:alg-passk}
\end{figure*}

In Figure \ref{fig:alg-passk}, we show the pass@$k$ results of models trained 
via improved algorithms. We find that the shaded region (range of improved 
algorithms) effectively collapses onto the GRPO curve (50\% noise), 
indicating that no algorithmic intervention successfully expands the set of 
solvable problems beyond the naive baseline. As sampling budget $k$ increases, 
the gap between these algorithms and the clean-data model (blue) widens, 
particularly on the rigorous AIME benchmark (Figure \ref{fig:alg-passk-aime}). 
This demonstrates that while these algorithms might stabilize training metrics, 
they fail to repair the fundamental damage noise inflicts on the model's ability 
to explore and find correct solutions.

\subsection{Generation Length Analysis of Different RLVR Algorithms}
\label{sec:app-len-algs}

\begin{figure*}[t!]
    \centering
    \begin{subfigure}{0.33\textwidth}
        \includegraphics[width=\linewidth]{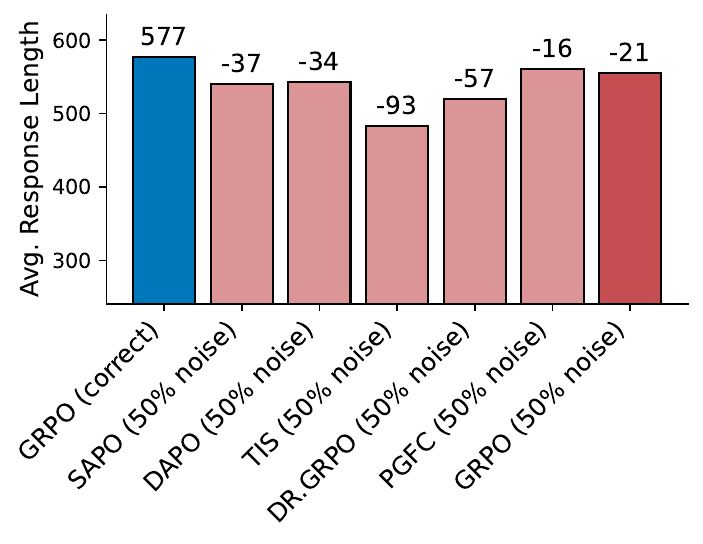}
        \caption{MATH-500.} 
        \label{fig:alg-length-math500}
    \end{subfigure}\hspace*{\fill}
    \begin{subfigure}{0.33\textwidth}
        \includegraphics[width=\linewidth]{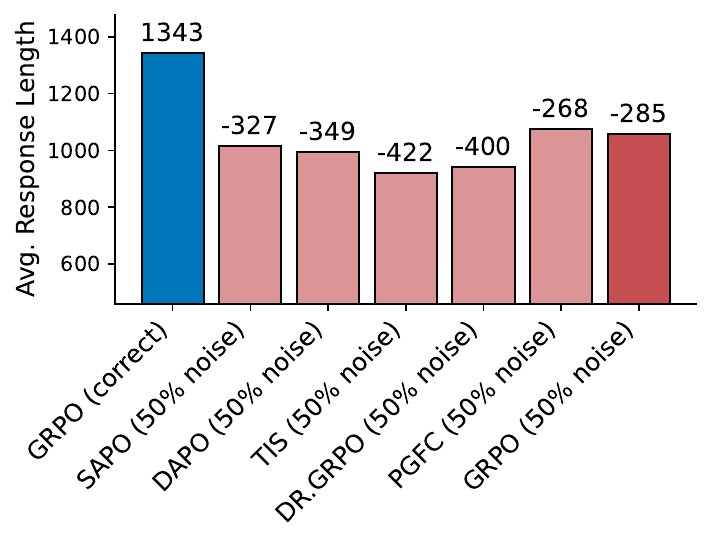}
        \caption{AIME 2024 and 2025.} 
        \label{fig:alg-length-aime}
    \end{subfigure}\hspace*{\fill}
    \begin{subfigure}{0.33\textwidth}
        \includegraphics[width=\linewidth]{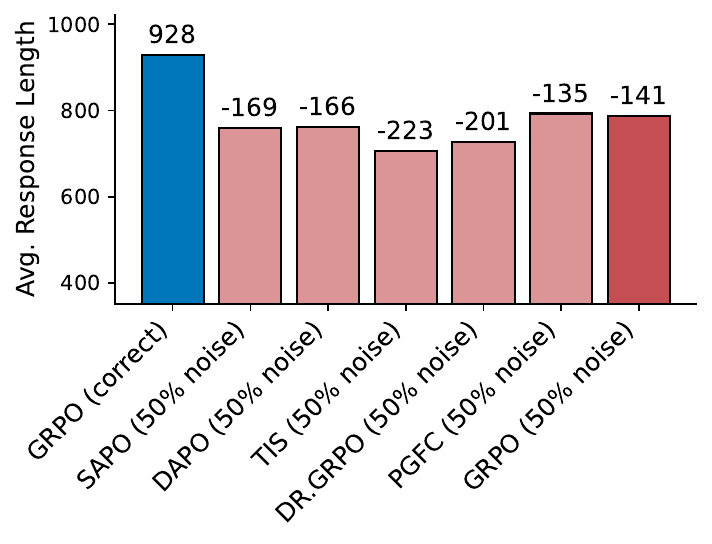}
        \caption{AMC 2023 and 2024.} 
        \label{fig:alg-length-amc}
    \end{subfigure}

    \caption{Training on noisy data with improved algorithms still results in 
    shorter responses.} 
    \label{fig:alg-length}
\end{figure*}

In Figure \ref{fig:alg-length}, we show the average response length across 
all trained models. We demonstrate a significant contraction in reasoning depth 
when noise is introduced. On the complex AIME benchmark (Figure 
\ref{fig:alg-length-aime}), the model trained on clean data generates long,
 detailed chains of thought (avg. 1343 tokens). In contrast, models trained on 
 noisy data, regardless of the algorithm used, exhibits a massive reduction in 
 length, dropping by over 300 tokens. This suggests that noisy rewards penalize 
 complex exploration, encouraging the model to converge on shorter reasoning 
 paths. Specifically, advanced algorithms like TIS and DAPO fail to reverse this 
 trend, often producing responses even shorter than the GRPO baseline. 
 This confirms that current algorithmic improvements do not restore the depth of 
 thought characteristic of models trained on high-quality data.

\section{Prompts Used in the Re-verification Pipeline}
\label{sec:app-prompt}

\newtcolorbox{mypromptbox}{
    colback=gray!15,       
    colframe=gray!15,      
    arc=5pt,               
    boxrule=1pt,           
    left=2pt,             
    right=2pt,            
    top=2pt,               
    bottom=2pt,            
    fontupper=\small\ttfamily,    
    halign=justify         
}

We used the following prompt when acquiring annotations from GPT-5 Pro. 

\begin{mypromptbox}
Question: \{question\}

If the question have multiple distinct and correct answers, provide all the 
answers. Write the answers in LaTeX format. Do not include any explanations.

Answer(s):
\end{mypromptbox}

We used the following prompt as the final prompt for judging the correctness of 
answers with GPT-5.
\begin{mypromptbox}
You are a highly intelligent and accurate math grader. Given a math question, a 
ground truth answer, and a student's answer, determine whether the student's 
answer is correct. You must follow the following guidelines: \\

1. If the question has multiple distinct but correct answers, the student only 
needs to provide one of them to be considered correct. For example, if the 
ground truth answer is "(k in {{12, -12}})" and the student's answer is "11", 
the student's answer should be considered correct.\\

2. Ignore any format mistakes and only grade the mathematical meaning of the 
answer. For example, if the ground truth answer is "(11)\_6" and the student's 
answer is "11", the student's answer should be considered correct. Another 
example is if the ground truth answer is latex formatted and the student's 
answer is not latex formatted but is mathematically equivalent, the student's 
answer should be considered correct. Vice versa is also true. \\

3. If the student's answer is mathematically equivalent to the ground truth 
answer, it should be considered correct. \\

4. If the student's answer uses fractions, decimals, or different 
representations that are mathematically equivalent to the ground truth answer, 
it should be considered correct. For example, if the ground truth answer is 
"4$\backslash$sqrt\{2\}" and the student's answer is "5.656854249492381", 
the student's answer should be considered correct. \\

5. If the question is an Yes/No question and the ground truth answer is an exact 
solution, the student's answer is correct as long as they answers "Yes" or 
anything equivalent to "Yes". For example, for the question "Does there exist a 
fraction equivalent to \$$\backslash$frac\{7\}\{13\}\$ such that the difference 
between the denominator and the numerator is 24?", the ground truth answer is 
"$\backslash$frac\{28\}\{52\}" and the student's answer is "Yes", the student's 
answer should be considered correct. \\

6. If the student's answer is a simplification of the ground truth answer, it 
should be considered correct. For example, if the ground truth answer is "2 + 2" 
and the student's answer is "4", the student's answer should be considered 
correct. Vice versa is also true. \\

7. If the student's answer is mathematically inequivalent to the ground truth 
answer, it should be considered incorrect. For example, if the ground truth 
answer is "5" and the student's answer is "-5", the student's answer should be 
considered incorrect. \\

Question: \{question\}

Ground Truth: \{ground\_truth\}

Student Answer: \{student\_answer\}

Is the student's answer correct? Answer 'Yes' or 'No'.

Answer: 
\end{mypromptbox}

\section{Determination of Training Duration}
\label{sec:app-ca}

\begin{figure*}[t!]
    \centering
    \begin{subfigure}{0.33\textwidth}
        \includegraphics[width=\linewidth]{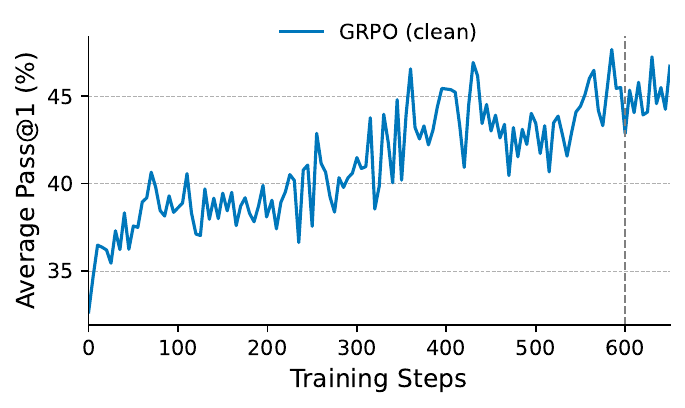}
        \label{fig:ca-clean}
    \end{subfigure}\hspace*{\fill}
    \begin{subfigure}{0.33\textwidth}
        \includegraphics[width=\linewidth]{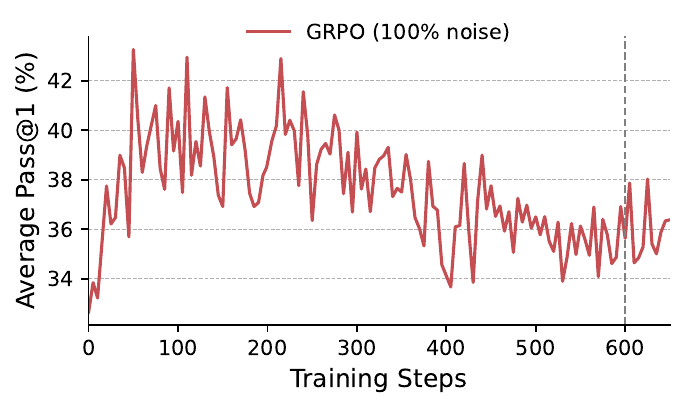}
        \label{fig:ca-format}
    \end{subfigure}\hspace*{\fill}
    \begin{subfigure}{0.33\textwidth}
        \includegraphics[width=\linewidth]{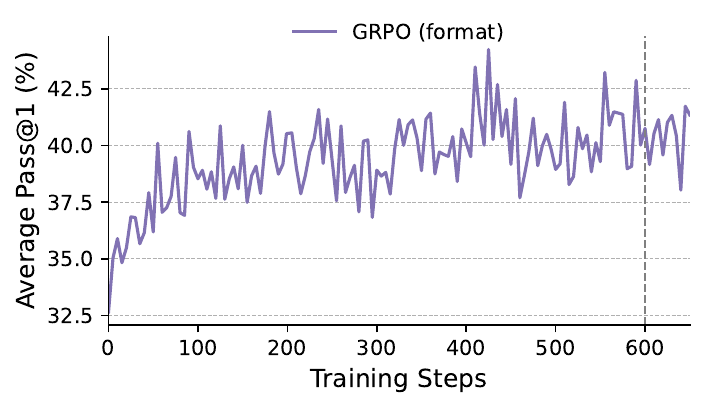}
        \label{fig:ca-noise}
    \end{subfigure}

    \caption{After three epochs (600 steps), no model shows improvements of
    average pass@1 of all benchmarks in the subsequent 50 steps.} 
    \label{fig:ca}
\end{figure*}

To establish a rigorous stopping criterion for math reasoning training, we 
monitored the average performance across all benchmarks over nearly 600 
training steps. As illustrated in Figure \ref{fig:ca}, the model trained on 
clean data (left) exhibits a steady ascent, reaching peak performance stability 
around 600 steps (3 epochs). In contrast, the model trained on 100\% noise (right) 
peaks early (approx. step 200) before degrading, confirming that prolonged 
exposure to noise leads to overfitting and performance collapse. The 
format-only model (center) stabilizes quickly and maintains a consistent pass@1. 
Consequently, we set the training duration to 600 steps for all experiments, as 
no model demonstrated significant improvements in the subsequent 50 steps.

\begin{figure*}[t!]
    \centering
    \begin{subfigure}[t]{0.33\textwidth}
        \includegraphics[width=\linewidth]{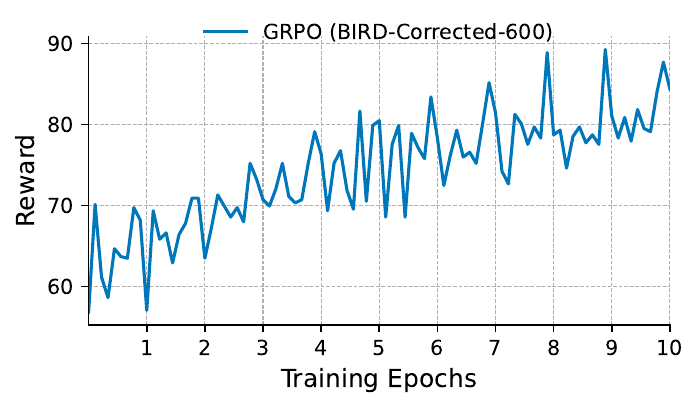}
    \end{subfigure}\hspace*{\fill}
    \begin{subfigure}[t]{0.33\textwidth}
        \includegraphics[width=\linewidth]{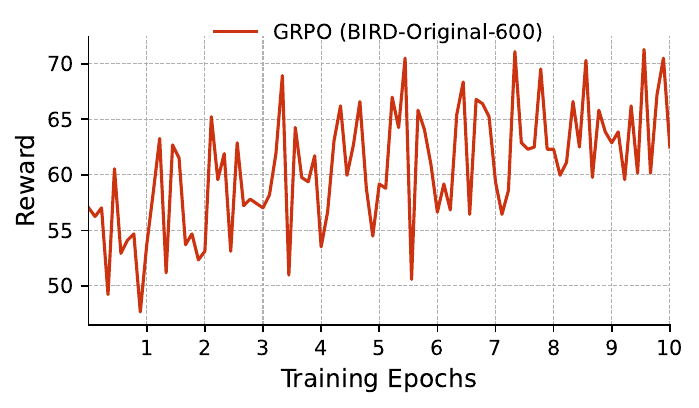}
        \caption{Qwen3-235B-A22B-Instruct-2507.} 
    \end{subfigure}\hspace*{\fill}
    \begin{subfigure}[t]{0.33\textwidth}
        \includegraphics[width=\linewidth]{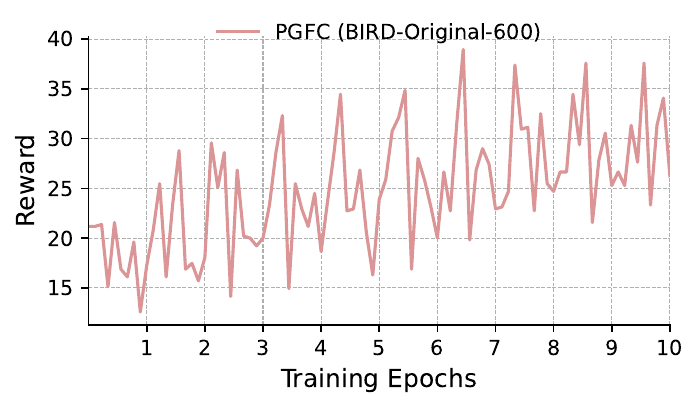}
    \end{subfigure}
    \vspace{1em}

    \begin{subfigure}[t]{0.33\textwidth}
        \includegraphics[width=\linewidth]{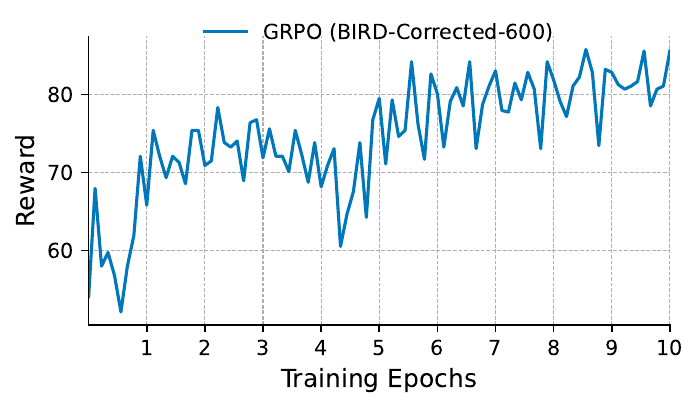}
    \end{subfigure}\hspace*{\fill}
    \begin{subfigure}[t]{0.33\textwidth}
        \includegraphics[width=\linewidth]{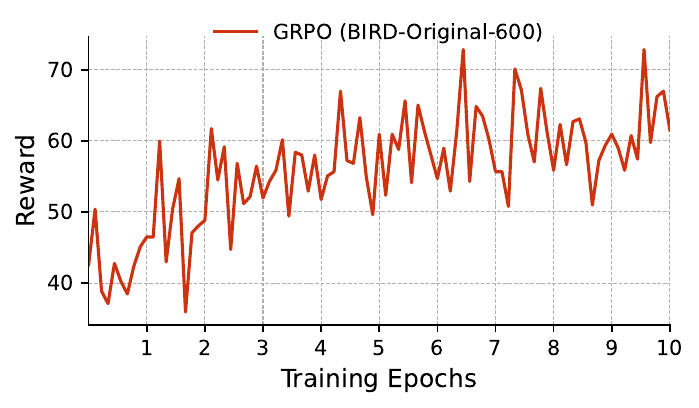}
        \caption{DeepSeek-V3.1.} 
    \end{subfigure}\hspace*{\fill}
    \begin{subfigure}[t]{0.33\textwidth}
        \includegraphics[width=\linewidth]{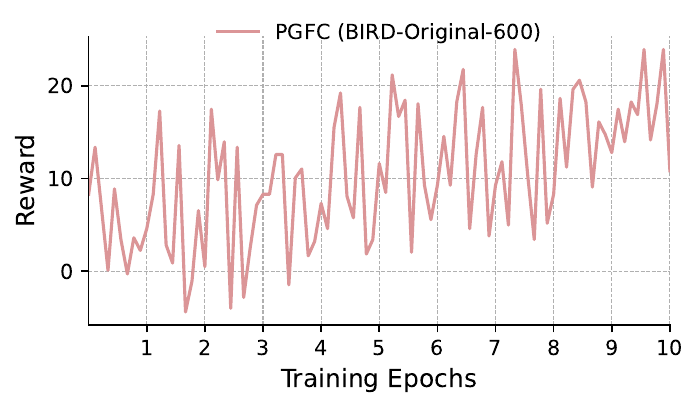}
    \end{subfigure}
    \vspace{1em}

    \begin{subfigure}[t]{0.33\textwidth}
        \includegraphics[width=\linewidth]{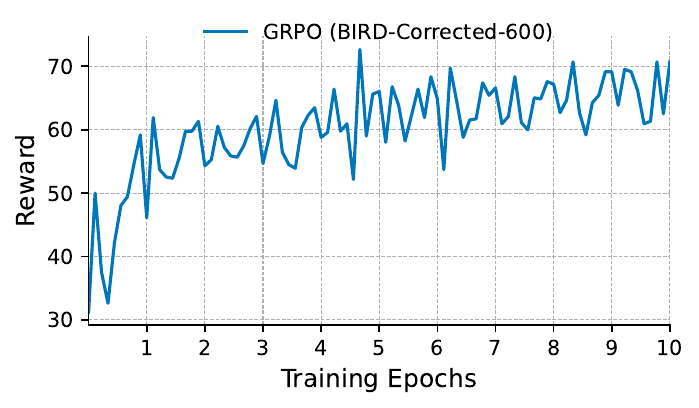}
    \end{subfigure}\hspace*{\fill}
    \begin{subfigure}[t]{0.33\textwidth}
        \includegraphics[width=\linewidth]{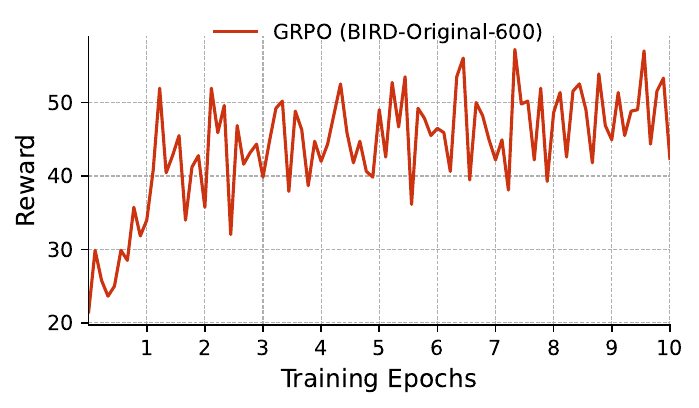}
        \caption{Qwen3-32B.} 
    \end{subfigure}\hspace*{\fill}
    \begin{subfigure}[t]{0.33\textwidth}
        \includegraphics[width=\linewidth]{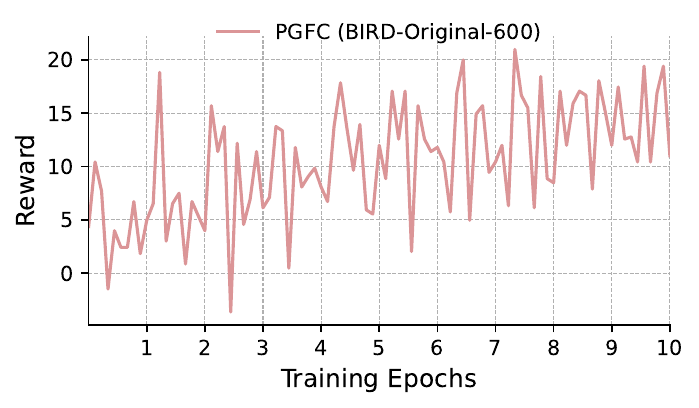}
    \end{subfigure}
    \vspace{1em}

    \begin{subfigure}[t]{0.33\textwidth}
        \includegraphics[width=\linewidth]{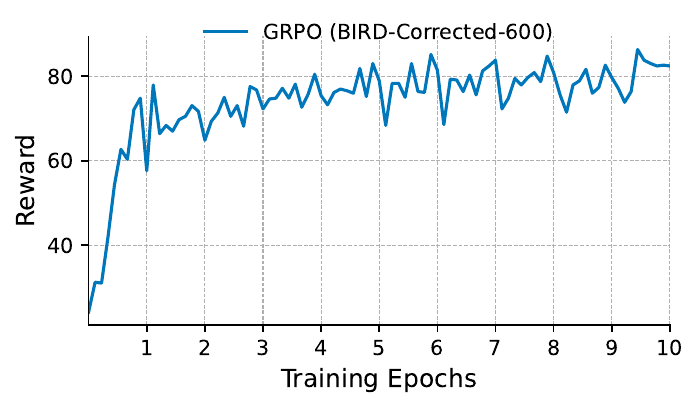}
    \end{subfigure}\hspace*{\fill}
    \begin{subfigure}[t]{0.33\textwidth}
        \includegraphics[width=\linewidth]{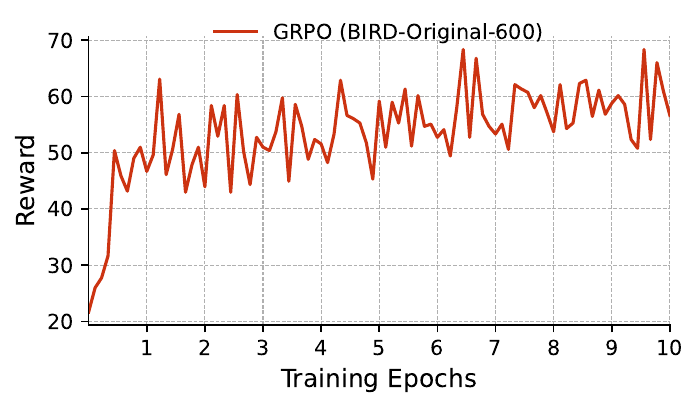}
        \caption{GPT-OSS-120B-A5.} 
    \end{subfigure}\hspace*{\fill}
    \begin{subfigure}[t]{0.33\textwidth}
        \includegraphics[width=\linewidth]{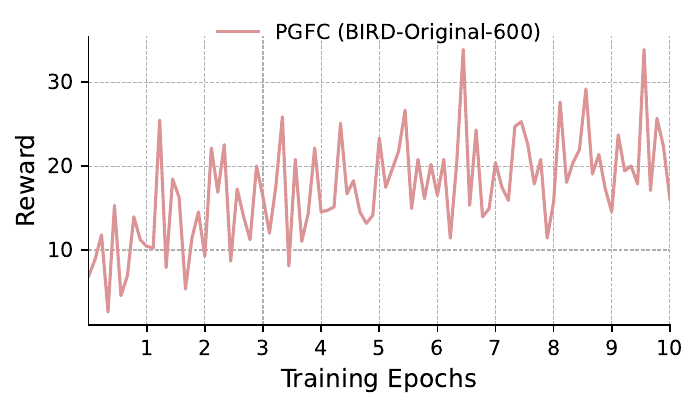}
    \end{subfigure}
    \vspace{1em}

    \begin{subfigure}[t]{0.33\textwidth}
        \includegraphics[width=\linewidth]{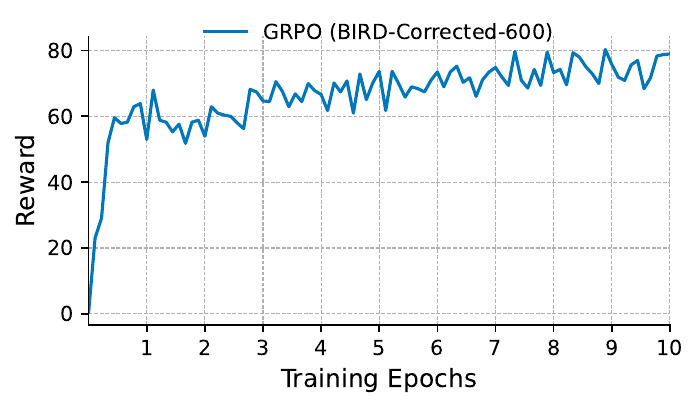}
    \end{subfigure}\hspace*{\fill}
    \begin{subfigure}[t]{0.33\textwidth}
        \includegraphics[width=\linewidth]{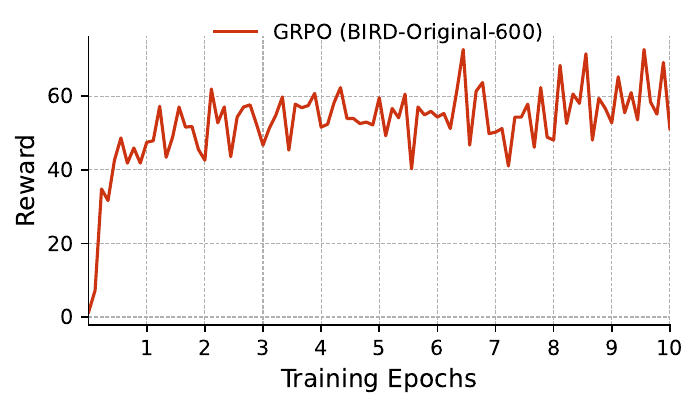}
        \caption{Llama-3.3-70B-Instruct.} 
    \end{subfigure}\hspace*{\fill}
    \begin{subfigure}[t]{0.33\textwidth}
        \includegraphics[width=\linewidth]{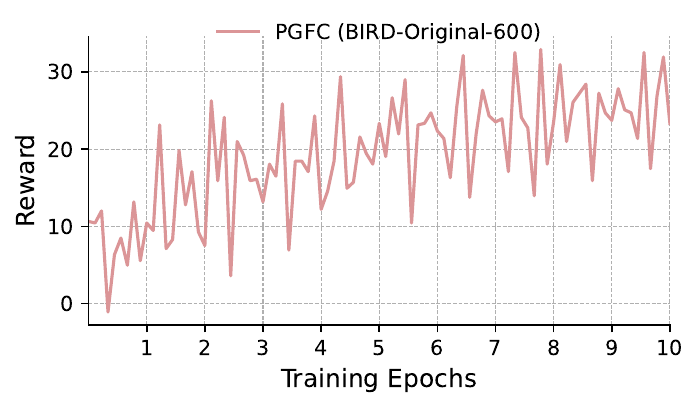}
    \end{subfigure}
    \vspace{1em}

    \caption{Starting the tenth epoch, no model shows improvements in terms of reward.}
    \label{fig:ca-bird}
\end{figure*}

Similarly, to determine the stopping criterion for Text2SQL training, we 
monitored the average training rewards. We did not monitor test accuracy on BIRD 
mini-Dev to prevent overfitting to a single benchmark. As illustrated in Figure 
\ref{fig:ca-bird}, starting the tenth epoch, no model shows improvements in 
training rewards. Consequently, we set the training duration to 10 epochs for 
all Text2SQL experiments.

\section{Training Hyperparameters}
\label{sec:app-hp}

We list all hyperparameters used in the math (Sections~\ref{sec:app-dyna}--\ref{sec:app-len-algs})
and Text2SQL experiments. Hyperparameters that differ across algorithms or
models are highlighted in their own tables; everything else is shared.

\subsection{Math RLVR Hyperparameters}
\label{sec:app-hp-math}

All math experiments fine-tune Qwen2.5-Math-7B with full-parameter updates on
the DeepScaleR training set under the SkyRL training stack. Settings shared by
every math run are given in Table~\ref{tab:math-common}. Algorithm-specific
deviations from this common configuration are listed in Table~\ref{tab:math-alg}.

\begin{table}[h]
    \centering
    \small
    \caption{Hyperparameters shared by every math RLVR run. All algorithms use
    the GRPO advantage estimator on top of these settings; algorithm-specific
    overrides are listed in Table~\ref{tab:math-alg}.}
    \label{tab:math-common}
    \begin{tabular}{@{}lll@{}}
        \toprule
        Category & Hyperparameter & Value \\
        \midrule
        Model & Base model & Qwen/Qwen2.5-Math-7B \\
              & Training strategy & FSDP2 (colocated train + generate) \\
              & GPUs per node & 4 \\
              & Gradient checkpointing & enabled \\
        \midrule
        Optimization & Train batch size (prompts) & 64 \\
              & Policy mini-batch size & 64 \\
              & Micro-batch per GPU (fwd / train) & 2 / 2 \\
              & Update epochs per batch & 1 \\
              & Max grad norm & 1.0 \\
              & KL loss coefficient & 0 \\
              & Learning rate & $5 \times 10^{-6}$ \\
              & Weight decay & 0.01 \\
        \midrule
        Rollouts & Samples per prompt (group size) & 16 \\
              & Max prompt length & 1024 tokens \\
              & Max generation length & 3072 tokens \\
              & Sampling temp.\ (train / eval) & 1.0 / 0.0 \\
              & vLLM GPU memory utilization & 0.8 \\
        \bottomrule
    \end{tabular}
\end{table}

\begin{table}[h]
    \centering
    \small
    \caption{Algorithm-specific hyperparameters. Entries marked ``--'' use the
    shared default from Table~\ref{tab:math-common}, including the learning
    rate of $5\times 10^{-6}$ and the zero KL coefficient. For algorithms
    inherited from prior work we follow the hyperparameters recommended in
    the original papers.}
    \label{tab:math-alg}
    \begin{tabular}{@{}lcl@{}}
        \toprule
        Algorithm & Loss aggregation & Other \\
        \midrule
        GRPO                         & token mean          & --- \\
        PGFC                         & token mean          & $\rho$ matched to dataset noise \\
        Dr.~GRPO                     & sequence mean & \texttt{grpo\_norm\_by\_std=false} \\
        DAPO                         & token mean      & see below \\
        TIS                          & token mean & DAPO + IS-ratio cap $2.0$ \\
        SAPO                         & token mean          & $\tau_{+}{=}1.0$, $\tau_{-}{=}1.05$ \\
        \bottomrule
    \end{tabular}
    \\[0.6em]
    \begin{minipage}{0.95\linewidth}
    \footnotesize
    \emph{DAPO settings:} $\epsilon_\text{low}{=}0.2$, $\epsilon_\text{high}{=}0.28$,
    $c{=}10.0$; dynamic sampling in \emph{filter} mode with at most $30$ sample
    batches; soft over-long penalty (buffer $2048$ tokens, penalty factor
    $1.0$); over-long filtering enabled. TIS inherits all DAPO settings on top
    of the importance-sampling ratio cap.
    \end{minipage}
\end{table}

\minihead{Loss aggregation.} The ``Loss aggregation'' column in
Table~\ref{tab:math-alg} controls how the per-token policy-gradient loss is
reduced into a single scalar per gradient step. Let $\ell_{i,t}$ denote the
token-level loss for the $t$-th token of the $i$-th rollout in the batch, and
let $|x_i|$ be the response length of rollout $i$. The two modes used in our
experiments are:
\begin{itemize}[leftmargin=1.2em,topsep=0pt,itemsep=2pt]
    \item \textbf{Token mean}:
    $\mathcal{L} = \tfrac{1}{\sum_i |x_i|} \sum_i \sum_{t=1}^{|x_i|} \ell_{i,t}$,
    i.e., a uniform average over every valid token in the batch. Long rollouts
    therefore contribute more to the gradient than short ones.
    \item \textbf{Sequence mean}:
    $\mathcal{L} = \tfrac{1}{B}\sum_{i=1}^{B} \tfrac{1}{T_{\max}} \sum_{t=1}^{|x_i|} \ell_{i,t}$,
    where $B$ is the batch size and $T_{\max}$ is the configured maximum
    generation length (3072 in our setup). Dividing each rollout's token-sum by
    the same constant $T_{\max}$ removes the length bias that token-mean
    aggregation introduces, as proposed by~\citet{liu2025understanding}.
\end{itemize}

\subsection{Text2SQL RLVR Hyperparameters}
\label{sec:app-hp-sql}

The Text2SQL experiments fine-tune five open-weight models with LoRA-style
adapters via the Tinker SDK against the BIRD training/test split described in
Section~\ref{sec:app-ca}. Because the training infrastructure (managed
distributed inference + LoRA fine-tuning) is identical across models, every
run uses the configuration in Table~\ref{tab:sql-common}; the only run-level
differences are the base model checkpoint, the training data file (clean
BIRD-Corrected vs.\ noisy BIRD-Original), and the PGFC noise rate where
applicable (Table~\ref{tab:sql-models}). All runs stop after 10 epochs, at
which point training reward has saturated for every model
(Figure~\ref{fig:ca-bird}).

\minihead{Importance-sampling policy loss.} The Text2SQL training loop in the
Tinker SDK separates the \emph{sampling policy} $q$ (the model that produced
the rollout, served from a managed inference fleet) from the \emph{learner
policy} $p_\theta$ (the on-device weights being updated). These two
distributions can drift apart between sampling and the gradient step because
rollouts are generated on a separate inference endpoint and inference engines
exhibit small numerical non-determinism relative to the trainer. To correct
the resulting off-policy bias, the SDK's \texttt{importance\_sampling} loss is
applied at the token level: each token's advantage is multiplied by the
learner-vs.-sampler probability ratio for that token, and the result is summed
along the sequence,
\begin{equation*}
    \mathcal{L}_{\text{IS}}(\theta)
    \;=\; -\sum_{t=1}^{|x|} \frac{p_\theta(x_t \mid x_{<t})}{q(x_t \mid x_{<t})}\, A(x_t)
    \;=\; -\sum_{t=1}^{|x|} \exp\!\big(\log p_\theta(x_t \mid x_{<t}) - \log q(x_t \mid x_{<t})\big)\, A(x_t),
    \label{eq:is-loss}
\end{equation*}
where $A(x_t)$ is the token-level advantage from group-relative reward
centering (the same GRPO-style estimator used in the math experiments),
$\log q(x_t \mid x_{<t})$ is recorded by the inference engine at sampling
time, and $\log p_\theta(x_t \mid x_{<t})$ is recomputed by the learner during
the forward pass. Unlike PPO, the ratio is \emph{not} clipped: this yields
the standard token-level REINFORCE-with-importance-sampling gradient and is
appropriate when the sampler and learner are expected to remain close (only
one optimizer step is taken per batch of rollouts in our setup). The
token-level loss is summed within each sequence and averaged across the
batch.

\begin{table}[h]
    \centering
    \small
    \caption{Hyperparameters shared by every Text2SQL RLVR run. Runs with PGFC
    additionally set the \texttt{noise\_rate} hyperparameter to the value in
    Table~\ref{tab:sql-models}; all other runs leave it unset.}
    \label{tab:sql-common}
    \begin{tabular}{lll}
        \toprule
        Category & Hyperparameter & Value \\
        \midrule
        Optimization & Batch size (prompts) & 64 \\
                     & Group size (rollouts per prompt) & 16 \\
                     & Learning rate & $5\times 10^{-5}$ \\
                     & LoRA rank & 32 \\
                     & Number of epochs & 10 \\
                     & Loss function & importance sampling \\
        \midrule
        Rollouts & Max input tokens & 32{,}768 \\
                 & Max output tokens per turn & 3{,}072 \\
                 & Per-rollout timeout & 180\,s \\
                 & Conversation prefix & enabled \\
                 & Curriculum learning & disabled \\
                 & Dynamic sampling & disabled \\
                 & Asynchronous mode & disabled \\
                 & Multi-turn iterative query refinement & enabled \\
        \midrule
        Evaluation & Sampling temperature & 0.0 (greedy) \\
                   & Test set & BIRD Mini-Dev (598 instances) \\
        \bottomrule
    \end{tabular}
\end{table}

\begin{table}[h]
    \centering
    \small
    \caption{Per-run differences for the Text2SQL experiments. The
    \emph{clean} condition trains on \texttt{bird-plat-588.parquet}
    (BIRD-Corrected) and the \emph{noisy} condition trains on
    \texttt{noisy-600.parquet} (BIRD-Original). PGFC uses the noisy training
    file together with the listed annotation-error-rate estimate.}
    \label{tab:sql-models}
    \begin{tabular}{lc}
        \toprule
        Model & PGFC noise rate $\rho$ \\
        \midrule
        Qwen3-235B-A22B-Instruct-2507 & 0.62 \\
        DeepSeek-V3.1                 & 0.62 \\
        Qwen3-32B                     & 0.62 \\
        GPT-OSS-120B-A5               & 0.62 \\
        Llama-3.3-70B-Instruct        & 0.62 \\
        \bottomrule
    \end{tabular}
\end{table}

\cleardoublepage


\end{document}